\documentclass[journal]{IEEEtran}
\usepackage{amsmath,amsfonts}
\usepackage{algorithmic}
\usepackage{algorithm}
\usepackage{array}
\usepackage[caption=false,font=normalsize,labelfont=sf,textfont=sf]{subfig}
\usepackage{textcomp}
\usepackage{url}
\usepackage{verbatim}
\usepackage{graphicx}
\usepackage{cite}
\usepackage{booktabs}
\usepackage{amsmath}
\usepackage{bbding}
\usepackage{epstopdf}
\usepackage{multirow}
\usepackage{fancyhdr}
\usepackage{xcolor}
\usepackage{pifont}
\usepackage[colorlinks,linkcolor=red]{hyperref}

\newcommand{\lisays}[1]{\textcolor{green}{[Li: #1]}}
\begin{document}

\title{Uncertainty-Induced Transferability Representation  for Source-Free Unsupervised Domain Adaptation}%\chensays{do we need the authorship list?}%\chensays{Trustworthy transfer: domain distributional representation uncertainty for source-free unsupervised domain adaptation}}

\author{Jiangbo~Pei\IEEEauthorrefmark{1},
Zhuqing~Jiang\IEEEauthorrefmark{1},
Aidong~Men,
Liang~Chen,
Yang~Liu,
Qingchao~Chen\textsuperscript{\ding{41}}
%\IEEEauthorrefmark{2}% <-this % stops a space
\thanks{Jiangbo Pei and Aidong Men are with the School of Artificial Intelligence, Beijing University of Posts and Telecommunications, Beijing 100876, China. Jiangbo Pei is also affiliated with the National Institute of Health Data Science, Peking University. (e-mail: jiangbop@bupt.edu.cn; menad@bupt.edu.cn).}% <-this % stops a space
\thanks{Zhuqing Jiang is with Beijing Key Laboratory of Network System and Network Culture, and also with Beijing University of
Posts and Telecommunications, Beijing 100876, China,(e-mail: jiangzhuqing@bupt.edu.cn).}% 
\thanks{Liang Chen is with School of Mathematical Sciences, Peking University, Beijing 100871, China,(e-mail: clandzyy@pku.edu.cn).}%
\thanks{Yang Liu is with Wangxuan Institute of Computer Technology at Peking University,  Beijing, 100080, China, (email: yangliu@pku.edu.cn).}% 
\thanks{Qingchao Chen is with the National Institute of Health Data Science, Peking University, Beijing, 100191, China. (e-mail: qingchao.chen@pku.edu.cn).}% 
\thanks{This work is supported by Peking University Medicine Seed Fund for Interdisciplinary Research (BMU2022MX011), the Fundamental Research Funds for the Central Universities and PKU-OPPO Innovation Fund BO202103.}%
\thanks{\IEEEauthorrefmark{1} Equally contributed first author.}%
\thanks{\ding{41} Corresponding author.}}% <-this % stops a 

\maketitle
\begin{abstract}
Source-free unsupervised domain adaptation (SFUDA) aims to learn a target domain model using unlabeled target data and the knowledge of a well-trained source domain model. 
Most previous SFUDA works focus on inferring semantics of target data based on the source knowledge.
Without measuring the transferability of the source knowledge, these methods \textit{insufficiently} exploit the source knowledge, and \textit{fail to} identify the reliability of the inferred target semantics.
However, existing transferability measurements require either source data or target labels, which are infeasible in SFUDA.
To this end, \textit{firstly}, we propose a novel Uncertainty-induced Transferability Representation (UTR), which leverages uncertainty as the tool to analyse the channel-wise transferability of the source encoder in the absence of the source data and target labels.
The domain-level UTR unravels how transferable the encoder channels are to the target domain and the instance-level UTR characterizes the reliability of the inferred target semantics.
\textit{Secondly}, based on the UTR, we propose a novel Calibrated Adaption Framework (CAF) for SFUDA, including i) \textit{the source knowledge calibration module} that guides the target model to learn the transferable source knowledge and discard the non-transferable one, and ii) \textit{the target semantics calibration module} that calibrates the unreliable semantics.
With the help of the calibrated source knowledge and the target semantics, the model adapts to the target domain safely and ultimately better. 
We verified the effectiveness of our method using experimental results and demonstrated that the proposed method achieves state-of-the-art performances on the three SFUDA benchmarks. Code is available at \url{https://github.com/SPIresearch/UTR}.
\end{abstract}

\vspace{-2mm}
\section{Introduction}
\begin{figure}
\centering
  \includegraphics[width=0.5\textwidth]{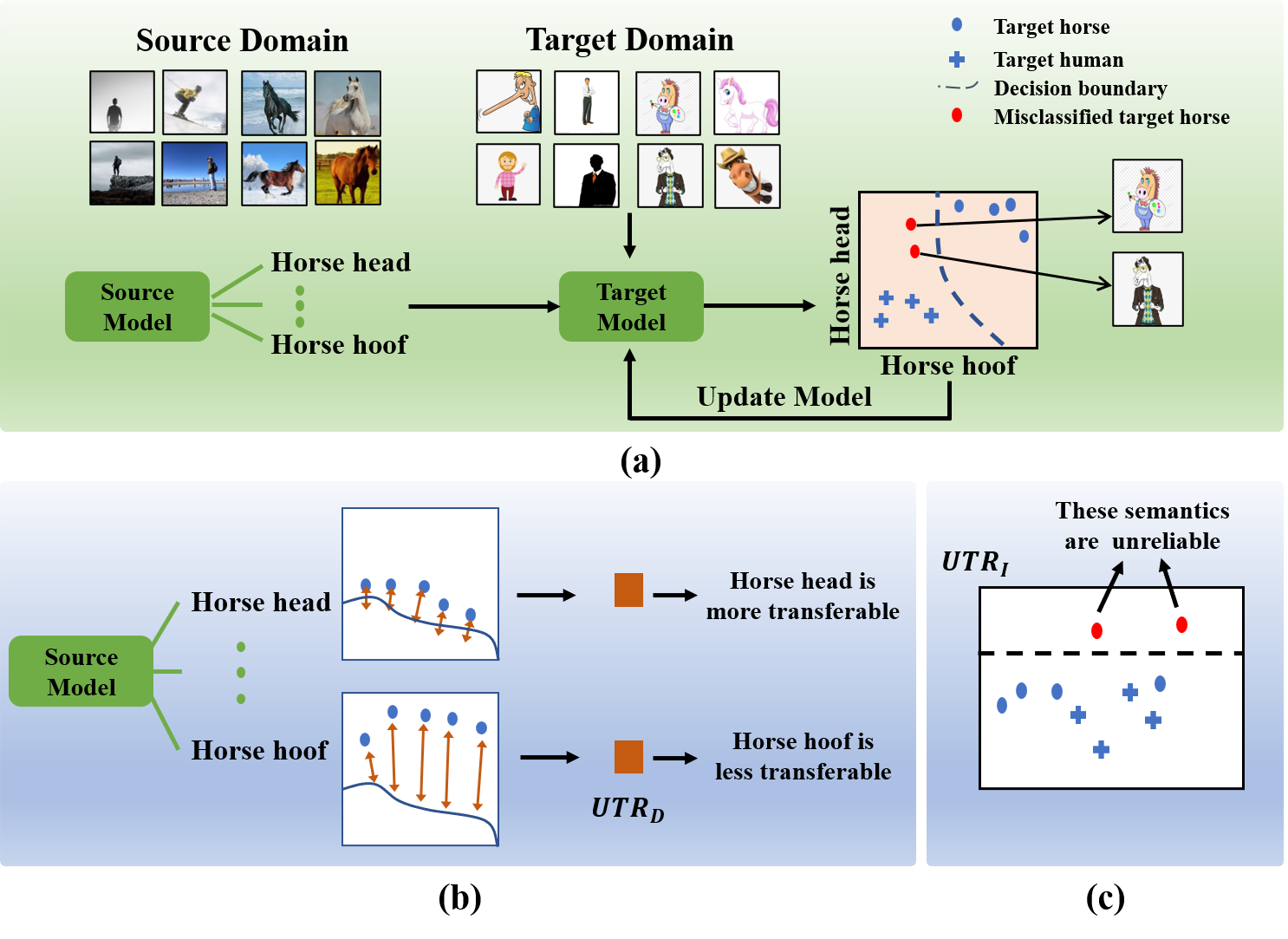}
  \vspace{-5mm}
  \caption{(a): Most existing SFUDA methods directly transfer all source knowledge to the target model at the start of training, infer the semantics (labels) of target data using the model, and update the model with the inferred semantics. Without identifying the source knowledge's transferability, the target model receives less-transferable knowledge (for example, the feature ``Horse hoof'' which is learned to classify humans and horses in the real-world source domain but may not be suitable for the target cartoon domain). The less-transferable knowledge hinders the model to infer the semantics of the target data (e.g. Misclassified Horse Image). (b): The $UTR_D$  identifies how transferable each channel of the source encoder is to the target domain.
 (c):  The $UTR_I$  characterizes the reliability of the inferred semantics of each target sample.}
  \vspace{-3mm}
  \label{fig:teaser}
\end{figure}

Deep neural networks have achieved state-of-the-art performance in a variety of image processing and computer vision applications when the testing data and training data are drawn from the same distribution (domain). When the model needs to be deployed in a new target domain (e.g. a new user uploads photos to a social media website), the recommendation or retrieval model often suffers from huge performance degradation due to the cross-user domain gap.
Unsupervised domain adaptation (UDA) is an effective solution to tackle the domain gap, which aims at adapting a model to a target domain where labels are not available with the help of a labeled source domain dataset. However, the vanilla UDA assumes the source data is accessible during adaptation, which is not always practical. On the one hand, data privacy protection is increasingly important because data often contain personal information. Sharing source domain data will endanger personal privacy and is strictly prohibited in many applications, especially in social media, medicine and biometrics. On the other hand, transmitting source data is costly such as video data or high-resolution images. 

Source-free unsupervised domain adaptation (SFUDA) is proposed as a promising task to tackle previous issues. SFUDA aims to learn a discriminative target domain model, given \textit{the unlabeled target domain data and a pre-trained source model but without any source data or labels}. 
To address SFUDA, as shown in Fig. \ref{fig:teaser} (a), most existing works \cite{yang2021exploiting,yang2021generalized,xia2021adaptive,yang2020unsupervised,liang2020we} directly transfer \textit{all} source knowledge to the target model at the start of training, infer the semantics (labels) of target data using the model and turn back to update the model with the inferred semantics.
However, these methods suffer two limitations. \textbf{\textit{Firstly, the utilization of the source knowledge is limited.}}
On the one hand, the way that transfers all source knowledge to the target model \textit{ignores discarding the non-transferable knowledge}. On the other hand, they transfer source knowledge at the start of training only, which wastes the valuable transferable knowledge learned from the well-annotated source domain.
\textbf{\textit{Secondly, the semantics of the target data inferred by the source knowledge is risky,}} if using the non-transferable knowledge (taking the ``human hoof'' in Fig. \ref{fig:teaser} (a) as an example). Updating the model using these risky semantics rarely learns a discriminative target model.

Therefore, the key common challenge and the missing part of existing SFUDA methods is to \textbf{\textit{measure the transferability of the source knowledge to the target domain in the absence of source data and target labels.}} To our best knowledge, only Wang et al. \cite{wang2022exploring} proposed to search for domain-invariant/transferable model parameters. They explore the transferability of source model parameters based on calculating their variations \textit{after} each adaptation procedure in the stochastic optimization. However, their measurement is susceptible to the quality of the adaptive procedures.

Beyond SFUDA, the transferability of the deep neural network has been studied intensively \cite{nguyen2020leep,you2021logme,ben2010theory,gretton2006kernel}. 

\textbf{\textit{Nevertheless, existing transferability  measurements require either source data or target labels, which are not applicable to SFUDA. }}

To tackle the key challenge, \textbf{\textit{we proposed a novel Uncertainty-induced Transferability Representation (UTR), which provides a transferability measurement to the source knowledge in the absence of source data and target labels}}.
Specifically, we develop the uncertainty as a tool to measure the transferability of the source model, inspired by the theory of distributional uncertainty \cite{gawlikowski2021survey,nandy2020towards,gao2020reducing} that measures how ``unfamiliar'' a trained model is with any input data, and the model uncertainty \cite{malinin2018predictive,gawlikowski2021survey,nandy2020towards} that reflects the degree to which the model fits its training distribution. 
Intuitively, the two uncertainties reveal the probability of the input data that are sampled from the training distribution of the model--that is, an implicit Uncertainty Distance (UD) between the input data and the training one. This measurement provides us solid theoretical supports and more importantly, we propose to bridge the uncertainty and the source model transferability in the SFUDA setup: \textit{ if a lower UD of the target data and source model is obtained, we made the following conjectures: the target data is ``closer'' to the source domain distribution in the encoding space of the model, which indicates the source knowledge (source model parameters) can more effectively eliminate the domain discrepancy between the two domains, reflecting the source knowledge is more transferable to the target domain.}
On the other hand, stemming from our finding that different channels of the source features have different transferability to the target domain, we propose to measure the transferability of the source encoder (the feature encoder of the source model) channel-wisely. Intuitively, the transferability of different channels reflects the transferability of the ``partial'' source knowledge that encodes the features in these channels, facilitating us to explore which ``partial'' source knowledge is transferable or non-transferable.

Our UTR can be considered as a transferability spectrum, consisting of the instance and channel axis, where the instance axis denotes which target data is used to calculate the UD for the transferability measuring, while the channel axis represents the transferability of different channels of the source encoder.
To facilitate the UTR to address the previous two limitations in SFUDA, we designed the following variants.
Specifically, for the first limitation, the UTR on the domain-level, namely $UTR_D$, integrates the UTR over all of the target instances, which measures the transferability of different channels more accurately than the UTR of each target instance, thus can efficiently guide the utilization of the knowledge of the source encoder.
For the second limitation, the instance-level namely $UTR_I$ integrates UTR of a particular instance over all channels, which is proven to characterize the reliability of the inferred target semantics of each target instance. The usages of the $UTR_D$ and $UTR_I$ are illustrated in Fig. \ref{fig:teaser} (b) and (c).

\textbf{\textit{Based on the introduced domain-level and instance-level UTR, we proposed a novel Calibrated Adaptation Framework to address the two limitations of existing SFUDA works.}} \textbf{\textit{Firstly,}} a source knowledge calibration module is designed, which uses $UTR_D$ to identify the transferability of different channels of the source encoder, and calibrates the source knowledge that transferred to the target domain by distilling the knowledge in transferable channels and discard the knowledge in less-transferable ones. \textbf{\textit{Secondly,}} a target semantic calibration module is proposed based on our $UTR_I$ to detect unreliable target semantics and calibrate them by designing a semantic calibration loss. The semantic calibration loss encourages the model to ``forget'' the unreliable semantics and ``discover'' the true ones. 
With the calibrated source knowledge and target semantics, we safely adapt the model to the target domain, therefore summarizing a better-performing target model.

Our main contributions are summarized as follows: \textbf{Firstly}, we propose an \textit{Uncertainty-induced Transferability Representation} (UTR) to explore the source model transferability in the absence of source data and target labels, which is beneficial to the SFUDA community. \textbf{Secondly}, we design a novel Calibrated Adaptation Framework (CAF) to calibrate the source knowledge and the inferred target semantics, allowing the target model to fully and safely exploit the source knowledge and target data, hence learning a better-performing target model.
\textbf{Finally}, we verified the effectiveness of our method with extensive experimental results and demonstrated that the proposed method achieves state-of-the-art performances on the three SFUDA benchmarks.

\vspace{-3mm}
\section{Related Work}

\subsection{Source free unsupervised domain adaptation}
Recent years have witnessed great achievements in the vanilla UDA \cite{ben2010theory,gretton2006kernel,long2017conditional,ganin2016domain,chhabra2021glocal,chhabra2021iterative,han2022learning,moon2022multi,deng2022dynamic,deng2021joint,xu2021neutral}.
However, they assume that the source data is accessible during the adaptation, which is not always practical.
SFUDA aims to adapt a source-trained model to an unlabeled target domain without access to source data \cite{liang2020we, ye2021source,yang2021model}. 
Without the labeled source data,  some methods propose to generate labeled data by generative adversarial net (GAN) \cite{goodfellow2014generative}. 
Kurmi et al.\cite{kurmi2021domain} generate source data using the source-trained classifier, so that the vanilla UDA methods can be applied. 
Li et al.\cite{li2020model} leverage a conditional GAN to directly produce training samples in the target style. These methods use the source model as auxiliary supervisions to control the label of the generated data. \textit{Nevertheless,} the source model is ineffective for the data generation process, due to the instability training of GAN \cite{arjovsky2017towards}.
Most existing SFUDA methods directly transfer all the source knowledge to the target model at the start of training, infer the semantic information of target data using the target model, and update the target model with the inferred semantic information.
SHOT \cite{liang2020we} and  ISFDA \cite{li2021imbalanced} predict the target category using the pseudo-labeling strategy. 
CPGA \cite{qiu2021source} and BAIT \cite{yang2020unsupervised} propose to align the samples with category-wise prototypes in a contrastive learning framework.
NRC\cite{yang2021exploiting} and LSC-SDA\cite{yang2021generalized} aim at propagating the categorical semantics from the neighborhood/cluster structure to the feature space.
Xia et al. \cite{xia2021adaptive} focus on the disagreements between target data and the source model. They select partial target data with high agreements with the source model and apply the source model to these samples.
\textbf{However, without measuring the transferability of the source knowledge, these methods fail to control over discarding non-transferable knowledge and preserving transferable knowledge. Additionally, they fail to identify risks of applying the source model to infer target semantics.}
% To our best knowledge, only Wang et al. \cite{wang2022exploring} relate to our works, which aim to explore the domain-invariant (i.e. transferable) parameters of the source model. They judge whether a parameter in a location is transferable by observing the value of parameters in the location of the source and target models.
% However, as they mentioned \cite{wang2022exploring}, the existence of the domain-invariant parameters is not proved, because the knowledge in a model is highly abstract and entangled and analyzing individual parameter generally makes less sense. 
To our best knowledge, Wang et al. \cite{wang2022exploring} is the most similar work as ours. They explored to transfer only partial source model parameters based on calculating the parameter variations \textit{after} each adaptation procedure
in the stochastic optimization. However, their measurement is susceptible to the quality of the adaptive procedures.
%On the other hand, although boosting the performance, as mentioned in \cite{wang2022exploring}, the exact ``domain-invariant'' parameters may not exist, because the knowledge in a model is highly abstract and entangled that hinders the analysis of transferability of individual parameters. 
\textbf{In contrast, our method measures the transferability using only the source model and unlabeled target data, which is irrelevant to the adaptation procedure, so that can avoid the hazards of the unreliable adaptation.} 
% However, as they mentioned \cite{wang2022exploring}, the existence of the domain-invariant parameters is not proved, because the knowledge in a model is highly abstract and entangled and analyzing individual parameter generally makes less sense. 

% \textbf{Differently,} we used the source features as the intermediary and can directly estimate the transferability of source features \textit{before} the adaptation procedure. On the one hand, our method can assess transferability more efficiently and intuitively than \cite{wang2022exploring}. On the other hand, our method naturally explores the transferability of the relevant source parameters that encode these features, which is verifiable by measuring the transferability of different source features.}

% availability source knowledge of the source knowledge. However, they aim to identify which target samples fit the source model and only apply the source knowledge in these samples. \textbf{Instead, we aim to measure the source knowledge transferability, i.e., what the source knowledge is transferable to the target domain.}  Compared with \cite{xia2021adaptive}, our method is more efficient to exploit the source knowledge, thus learning a more discriminative target model.

\begin{figure*}[t]
  \centering
  
   \includegraphics[width=0.7\linewidth]{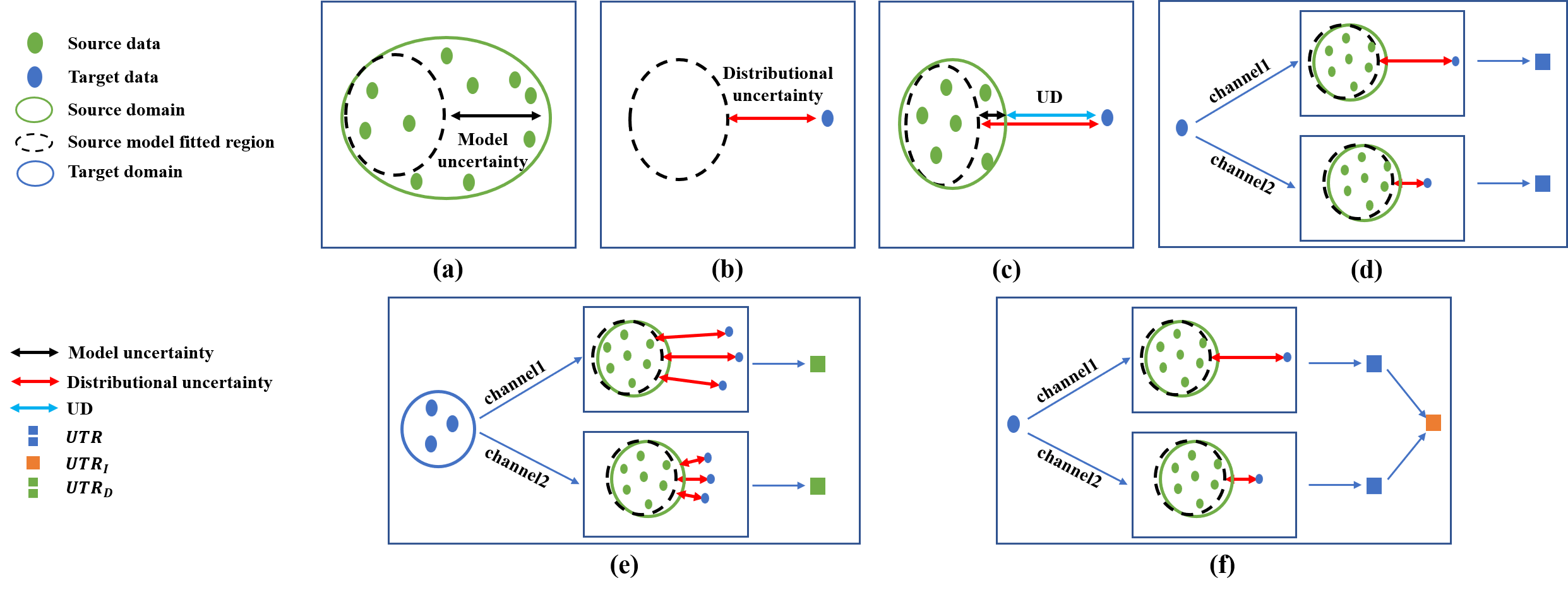}
   \vspace{-3mm}
   \caption{(a): The model uncertainty measures the degree to which a model's fitted region covers its training distribution. (b) The distributional uncertainty measures  the probability of an input instance that is sampled from a region that unfitted/unfamiliar by the model, which reveals how far the sample  is from the fitted region of the model.
   (c):  The distributional and model uncertainties reveal an implicit uncertainty distance (UD) from the target instance to the source data distribution, which reveals the ability of the source model in reducing the domain discrepancy between the target and source domains, therefore suggesting the transferability of the source model to the target domain. In SFUDA, UD could be approximated by distributional uncertainty as the model uncertainty is small. 
   (d) The UTR leverages the distributional uncertainty to estimate transferability of different channels of the source encoder to the target domain.
   (e) The domain-level UTR integrates the UTR over all target instances to estimate the transferability of these channels more accurately.
   (f) The instance-level UTR integrates UTR on the channel axis, which identifies the risk of using source knowledge to predict the semantics of the target instance.}
   \vspace{-6mm}
   \label{fig:unc_illu}
\end{figure*}

\vspace{-4mm}
\subsection{Uncertainty}
\vspace{-1mm}
Uncertainty is an important criterion to measure the robustness of a deep model\cite{kwon2021repurposing,kendall2017uncertainties,gawlikowski2021survey,sensoy2018evidential}.
Given an annotated sample $(x,y)$ and a model parameterized by $\theta$ trained on domain $D$, the uncertainty can be decomposed into the following equation:
{\setlength\abovedisplayskip{1pt}
\setlength\belowdisplayskip{1pt}
\begin{equation}
P(y|x,D)= \iint  \underbrace{P(y|\mu)}_{Data} \underbrace {P(\mu|x,\theta)}_{Distributional} \underbrace {P(\theta|D)}_{Model} d\theta d\mu,
\label{eqa:uncertainty}
\end{equation}}where $\mu=\theta(x)$ is the predicted label distribution and the three probability density functions represent the data uncertainty, model uncertainty, and distributional uncertainty respectively \cite{gawlikowski2021survey,nandy2020towards,gao2020reducing}.
The data uncertainty is almost irreducible which arises from the natural complexity of the data, such as the class overlap, label noise, homoscedastic and heteroscedastic noise. The model uncertainty measures how well the model fits to its training distribution \cite{malinin2018predictive,gawlikowski2021survey,nandy2020towards}. 
% In other words, it characterizes the degree to which the fitted region of the model covers its training distribution.
The distributional uncertainty measures the probability of an input instance that is sampled from a region that the model is ``unfamiliar'' with. 
Its characteristic has prompted its usage in the out-of-distribution detection \cite{sedlmeier2019uncertainty,padhy2020revisiting,mcallister2019robustness} and also in vanilla UDA methods \cite{gao2020reducing,liang2019exploring}. 
\textbf{To our best knowledge, our work is the first to propose the use of uncertainty to explore transferability in SFUDA.}

%\peisays{need to modify}
%\textbf{The proposed DDRU is different from the existing distributional uncertainty as follows:}
%\textit{Firstly}, the previous distributional uncertainty methods\cite{gawlikowski2021survey,nandy2020towards,gao2020reducing,gao2020reducing} evaluate the whole model, however, our DDRU estimates the transferability of the feature representations within the source model. DDRU explores what knowledge is transferable to the target domain, rather than the whole source model. \textit{Secondly,} the previous distributional uncertainty represents the uncertainty of an instance about the training distribution, while our DDRU captures the uncertainty of the whole target domain facing a source model to adapt.

\vspace{-5mm}
\subsection{Transferability}
\vspace{-2mm}
It is essential to asses and measure the model transferability and there are two mainstream methods in the deep learning community.
\textbf{Firstly,} the transferability of a model is measured by how much it can bridge the domain discrepancy between the source and the target domain\cite{chen2019transferability}. It can be calculated by domain discrepancy measurements such as Proxy $\mathcal{A}$-distance \cite{ben2010theory} and Maximum Mean Discrepancy (MMD) \cite{gretton2006kernel}.
In addition, Chen et al. \cite{chen2019transferability} propose the Corresponding Angle to measure the transferability.
\textbf{However, these methods require the access to the source data which is not suitable for SFUDA.}
\textbf{Secondly,} some transfer learning methods investigated the transferability of pre-trained source representations to the target domain. 
Existing works in this line of research have been proposed, such as the NCE \cite{tran2019transferability}, LEEP \cite{nguyen2020leep} and LogME \cite{you2021logme}.
\textbf{\textit{Nevertheless,} they need the target data annotations which are not applicable in SFUDA.}  
%In the SFUDA, the source data is inaccessible and the target label is unknown, which makes both the two manners infeasible.% Therefore, it is challenging to identify the source knowledge transferability in SFUDA.
\textbf{In contrast, our proposed method can estimate the transferability in the absence of source data and target data labels that fits the challenging SFUDA setup.}
\vspace{-3mm}
\section{Uncertainty-Induced Transferability Representation}
\vspace{-2mm}
It is essential to analyse the source knowledge transferability for SFUDA, \textit{however,} existing transferability measurements are not applicable in SFUDA. To tackle this problem, in Section \ref{sec::UD}, we develop the Uncertainty Distance (UD) as a tool to estimate the general transferability in the absence of source data and target annotations. In Section \ref{sec::UTR}, we introduce the channel-wise transferability analysis and propose the Uncertainty-induced Transferability Representation (UTR). In Section \ref{sec::Spectrum}, we derive the domain-level UTR and the instance-level UTR and state their effectiveness for the SFUDA community.
\label{sec::UTR_D}

\vspace{-4mm}
\subsection{Transferability measurement using Uncertainty Distance} 
\vspace{-2mm}
\label{sec::UD}
Not all knowledge in the source model is transferable and discriminative to the target domain. Therefore, it brings risks if we do not measure and quantify the transferability of source knowledge but deploy it directly in the target domain.
However, previous transferability measurements require some matched information, either both the source and target data, or data-annotation pairs of the target domain. These requirements are infeasible in SFUDA where only unmatched source model and target data are provided. 
The unmatched information makes it extremely challenging to measure the transferability by acquiring {\textit{``known and certain''}} information as the supervision signal. 

To this end, our work alternatively explores and exploits the uncertainty as a fundamental tool, and proposes an Uncertainty Distance (UD) to address these challenges. The UD is an implicit distance between the target instance $x_t$ and the source domain $D_s$. A low UD demonstrates that $x_t$ is ``close'' to $D_s$ given a source model parameterized by $\theta_s$, which reflects that it is efficient for the $\theta_s$ to reduce the domain discrepancy between the source and target domains. Therefore it suggests that the $\theta_s$ is transferable to the target domain and a high UD indicates the opposite.

Our consideration is shown in Fig. \ref{fig:unc_illu} (a)-(c). Given the source model parameterized by $\theta_s$, the model uncertainty characterizes the degree to which the fitted region of $\theta_s$ covers its training distribution (i.e. the source domain $D_s$). While given both the $\theta_s$ and the target instance $x_t$, the distributional uncertainty reveals how far the $x_t$ is from the fitted region of the $\theta_s$. Previous observations inspired us that \textit{the cooperation of the two uncertainties reveals the distance between the target instance $x_t$ and the source domain $D_s$}. Such a distance implicitly reflects the contributions of the source model to reduce the domain discrepancy. It can also be used to probe and measure the transferability of the source model to the target instance for SFUDA.

By incorporating the distributional uncertainty and the model uncertainty, we first formulate the UD as: 
{\setlength\abovedisplayskip{1pt}
\setlength\belowdisplayskip{1pt}
\begin{equation}
\begin{aligned}
UD(x_t,\theta_s,D_s)&= M( \underbrace {P(\theta_s(x_t)|x_t,\theta_s)}_{Distributional} \underbrace {P(\theta_s|D_s)}_{Model}),
\end{aligned}
\label{eqa:unc_cal_1}
\end{equation}}where $M(\cdot)$ is the uncertainty measurement function such as the Sensitivity Analysis \cite{nagy2007distributional}, the Deep Ensembles\cite{lakshminarayanan2016simple} and the MC dropout \cite{gal2016dropout}.

Although it requires the $D_s$ in Equation \ref{eqa:unc_cal_1} to measure the model uncertainty, we argue that \textit{it is still feasible to estimate transferability using UD in the SFUDA.} The reason is that the source model has been well-trained in the source domain so that the $\theta_s$ fits $D_s$ well. As shown in Fig. \ref{fig:unc_illu} (c), in this case, the model uncertainty is small enough to be ignored, and the UD in SFUDA can be approximated by the distributional uncertainty calculated by the target instance $x_t$ and source model parameters $\theta_s$:
{\setlength\abovedisplayskip{1pt}
\setlength\belowdisplayskip{1pt}
\begin{equation}
UD(x_t,\theta_s)= M({P(\theta_s(x_t)|x_t,\theta_s)}).
\label{eqa:unc_cal_2}
\end{equation}}
\vspace{-9mm}
\subsection{Channel-wise Transferability Analysis} 
\vspace{-2mm}
\label{sec::UTR}
The proposed UD in Equation \ref{eqa:unc_cal_2} essentially measures the transferability of the \textbf{whole source knowledge} (i.e., the whole $\theta_s$) to the target instances. Nevertheless, as motivated in the introduction, only partial knowledge is useful for the target domain. Therefore we proposed to analyse the transferability of the knowledge in a finer-grained manner: to determine which part of the learned source parameters are transferable to the target domain. A straight-forward method is to measure the transferability of the partial and individual source parameters $\theta$ using $UD(x_t,\theta)$, where $\theta \subset \theta_s$.
% In the SFUDA, the source model is given and we are interested in analyzing the transferability of the knowledge (the learned parameters) within the source model, so as to determine what internal knowledge of the source model is transferable to the target domain.
% A natural idea is: for the $\theta$ belongs to the whole source parameters $\theta_s$, measuring its transferability by $UD(x_t,\theta)$.
\textbf{\textit{However,}} it is well-known that most deep neural networks belong to the end-to-end ``black-box'' system, where the knowledge is highly abstract and entangled. Individual parameters generally make no sense, let alone analyzing their transferability.
 
To tackle this challenge,  we propose to estimate the transferability of \textbf{\textit{different channels of the source encoder rather than different model parameters}}, as shown in Fig. \ref{fig:unc_illu} (d). In this way, the transferability of a particular channel natural represents the transferability of the "partial" source knowledge (relevant parameter) that encodes the feature of this channel.
% Following this strategy, we explore the transferability of the relevant source knowledge to encode these features. 
More specifically, we propose the Uncertainty-induced Transferability Representation (UTR), a transferability spectrum, composed of the instance axis and the channel axis, which is formulated as:
{\setlength\abovedisplayskip{1pt}
\setlength\belowdisplayskip{1pt}
\begin{equation}
\begin{aligned}
UTR(x_t,h_{s})=[UD(x_t,h_{s}^1),..., UD(x_t,h_{s}^d)],
\end{aligned}
\label{eqa:UTR}
\end{equation}}where $UD(x_t,h_s^i)=M(P(z_i|x_t,h_{s}^i))$, $x_t$ is the target instance, $z=h_s(x_t), \; z\in\mathcal{R}^{d}$ denotes the $d$-channel target features produced by the source encoder $h_s$, and $z_i=(h_s(x_t))^i=h_s^i(x_t)$ is the target feature of the $i^{th}$ channel, $h_s^i$ is the potential source parameter to encode $z_i$.
The instance axis of UTR denotes which target data is used to calculate the UD for the transferability estimating.
The channel axis represents the transferability of different channels of the source encoder.
The $i^{th}$ channel of the UTR (i.e., $UD(x_t,h_s^i)$) indicates the transferability of  $z_i$ to the target domain. A low value of $UD(x_t,h_{s}^i)$ indicates that the target instance $x_t$ is close to the source one in the space of $z_i$ and suggests that the source knowledge to encode $z_i$ (i.e., the parameters $h_s^i$) is highly transferable across the two domains.
% which demonstrates the effectiveness of the feature $i$ to reduce the domain discrepancy between the distribution behind $x_t$ (the target domain) and the source domain, and suggests the transferability of the relevant source model parameters (knowledge) to encode the feature.

% the target instance $x_t$ is close to the source domain in the feature $i$ space}, which demonstrates the effectiveness of the feature $i$ to reduce the domain discrepancy between the distribution behind $x_t$ (the target domain) and the source domain, and suggests the transferability of the relevant source model parameters (knowledge) to encode the feature. 

To calculate the UTR, we adopt the sensitivity analysis method \cite{nagy2007distributional} as the uncertainty measurement $M(\cdot)$ for Equation \ref{eqa:UTR}. To be specific, the model parameters of $h_s$ are perturbed for $T$ times randomly as follows: $\{h_{s;T}=(1+r_t)*h_{s}\}_{t=1}^T$ are \textit{firstly} calculated by inserting $T$ random perturbations $\{r_t\}_{t=1}^T$ to original parameter $\theta_{hs}$.
\textit{Then} the uncertainty is estimated by calculating the variance of the $T$ outputs of the $i^{th}$ dimension feature:
{\setlength\abovedisplayskip{1pt}
\setlength\belowdisplayskip{1pt}
\begin{equation}
\begin{aligned}
   M(P(z_i|x_t,h_{s}^i))= Var_{h_s \sim h_{s;T}}((h_{s}(x_t))^i).
\end{aligned}
  \label{eqa:mUTR_D}
\end{equation}}

\begin{figure*}[t]
  \centering
  
   \includegraphics[width=0.9\linewidth]{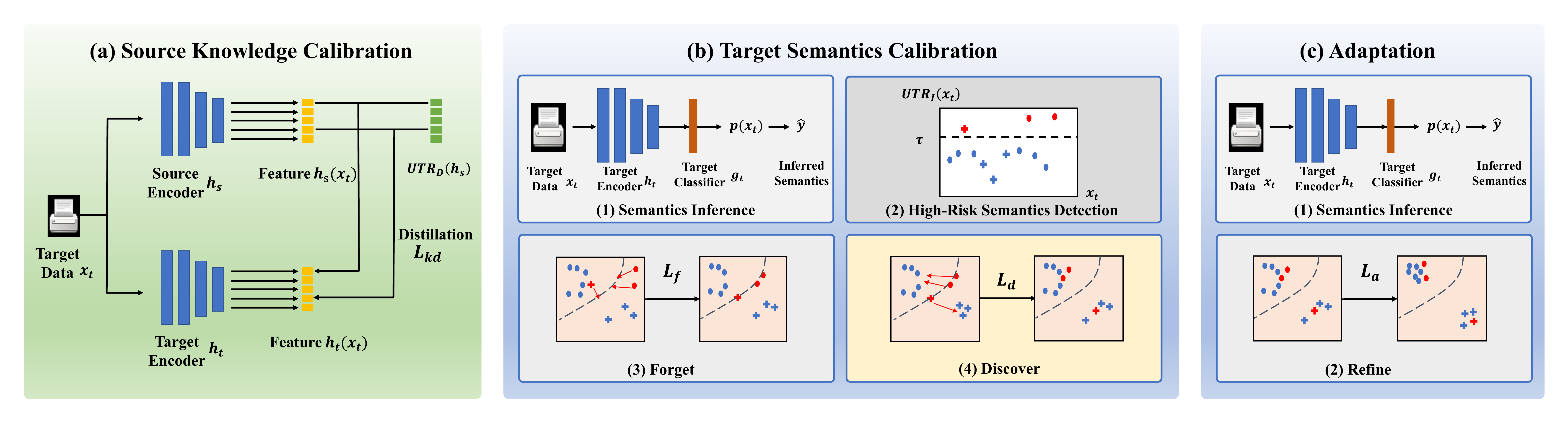}
   \vspace{-3mm}
   \caption{Overview of Our Calibrated Adaptation Framework. (a) Source knowledge absorption calibration. The $UTR_D$ is calculated and used to estimate the transferability of the knowledge of the source encoder $h_s$. Then knowledge in $h_s$ is distilled into the target encoder $h_t$ with $L_{kd}$, which controls the target encoder to absorb transferable source knowledge and neglect less-transferable knowledge according to $UTR_D$. 
   (b) Target semantics calibration. (b.1) Infer target semantics with target model.
   (b.2) Select instances whose inferred semantics are risk (the red point) according to their $UTR_I$ and threshold $\tau$.
   (b.3) The forget objective $L_{f}$ of the semantics calibrate loss minimizes the negative cross-entropy to risk instances, forcing \lisays{it} to forget the current unreliable semantics. (b.4) The discover objective $L_{d}$ of the semantics calibrate loss guides to discover their true semantics by minimize the entropy of the prediction probability distribution of the target instances. (c) Adaptation. (c.1) Re-infer  target semantics. (c.2) Refine the model with the adapt loss $L_{a}$.
 }
   \vspace{-5mm}
   \label{fig:method}
  
\end{figure*}
\vspace{-6mm}
\subsection{The Domain-level and Instance-level UTR}
\vspace{-2mm}
\label{sec::Spectrum}
Given a source model parameterized by $\theta_s$ and a target instance $x_t$, the UTR (in Equation \eqref{eqa:UTR}) is able to quantify the fine-grained transferability of the instance-level target features. In order to tackle the limitations in the SFUDA community: 1) measuring the transferability of source knowledge to target domain to sufficiently exploit it; 2) measuring the risk of inferring semantic information of target instances using the source knowledge, we design two variants of the UTR on two levels: the \textit{domain-level} UTR namely the \textit{$UTR_D$} and the \textit{instance-level} UTR namely the \textit{$UTR_I$}.

The $UTR_D$ describes the domain-level transferability estimation over the channel axis, which identifies how transferable each channel of the source encoder is to the target domain using the UD of the source model to all target instances. The $UTR_I$ characterizes the instance-level trasferability over all the target instances, which identifies the instance-level risk of inferring target semantic labels. The two are useful measurements proposed to fit in the later on adaptation framework for SFUDA problem. 

Specifically, as shown in Fig. \ref{fig:unc_illu} (e), the \textit{$UTR_D$} is calculated by integrating the $UTR(x_t,h_{s})$ of all $n_t$ target instances over the target domain $D_t$. Detailed formulation is as follows:  
{\setlength\abovedisplayskip{1pt}
\setlength\belowdisplayskip{1pt}
\begin{equation}
\begin{aligned}
UTR_D&(h_{s})=\mathbb{E}_{x_t \sim \mathcal{X}_t } UTR(x_t,h_{s})\\
&=\frac{1}{n_t}[\sum_{i=0}^{n_t} UD(x_t^i,h_{s}^1),..., \sum_{i=0}^{n_t} UD(x_t^i,h_{s}^d)]
\end{aligned}
\label{eqa:UTR_D}
\end{equation}}

As for the instance-level transferability spectrum, the $UTR_I$ is calculated by integrating the $UTR(x_t,h_{s})$ over all the $d$-channels of the source encoder $h_s$, as shown in Fig. \ref{fig:unc_illu} (f). The detailed formulation of the $UTR_I$ is as follows:
{\setlength\abovedisplayskip{1pt}
\setlength\belowdisplayskip{1pt}
\begin{equation}
\begin{aligned}
UTR_I(x_t)=\mathbb{E}_{z \sim \mathbb{R}^{d}} UTR(x_t,h_s)\\
=\frac{1}{d}[\sum_{i=0}^{d} UD(x_t,h_{s}^i)]
\end{aligned}
\label{eqa:UTR_I}
\end{equation}}

\vspace{-4mm}
\section{Calibrated Adaptation Framework}

\label{sec::ica}
\vspace{-2mm}
\subsection{Notation}
\vspace{-2mm}

In this paper, we focus on the $K$-way visual object classification task. 
SFUDA provides a well-trained source model parameterized by $\theta_s$ to the target domain $D_t$, where $\theta_s=g_{s} \circ h_{s}$, $h_{s}$ is the parameter of the source encoder, and $g_{s}$ is the parameter of the source classifier. The target domain $D_t=\{x_t^i\}_{i=1}^{n_t}$ consists of $n_t$ unlabeled target instances. The SFUDA aims to learn a discriminative target model parameterized by $\theta_t= g_{t} \circ h_{t}$ using the $\theta_s$ and $D_t$.
\vspace{-5mm}
\subsection{Overall}
\vspace{-2mm}
Most existing SFUDA methods directly transfer all source knowledge to the target model at the start of training, infer the semantic information (target labels) of target data using the model, and directly update the model using the inferred semantic information. \textit{However, they are limited as follows:} 1) the utilization of the source knowledge is limited. Directly transferring all source knowledge to the target model ignores discarding the less-transferable one. And updating the models using the inferred target semantics failed to preserve the discriminative knowledge in the source model. 2) the target semantic information inferred by the source model is risky due to the less-transferable source knowledge. Refining the model using the risky semantic information is unreliable.
% 1) directly transferring all source knowledge to the target model ignores discarding the non-transferable one, while rapidly updating the knowledge wastes the  part of transferable knowledge learned from fully-annotated source domain in a reliable supervised manner.  
% 2) due to the non-transferable knowledge, the semantic information inferred by the source knowledge is risky. Refining the model using the risky semantic information is unreliable.

To this end, we introduce the Calibrated Adaptation Framework (CAF). To tackle the first limitation, we propose to \textbf{\textit{calibrate the source knowledge that transferred to the target model}} using our $UTR_D$. To tackle the second, we propose to \textbf{\textit{calibrate the inferred semantic information of target instances}} based on our $UTR_I$. Finally, we adapt the model based on the calibrated source knowledge and target semantics. The overview of CAF is shown in Fig. \ref{fig:method}. The pseudo-code of the whole algorithm is described in Algorithm \ref{algorithm}.

% To tackle the two limitations, we introduce the Calibrated Adaptation Framework (CAF), which proposes to calibrate source knowledge absorption process of the target model based on our $UTR_D$ to address the challenge 1,  while calibrates the inferred semantic information of target instances by $UTR_I$ to address the challenge 2, finally adapt the model to the target  domain based on the calibrated knowledge and target semantics. The overview of CAF is shown in Figure \ref{fig:method}. The whole algorithm is illustrated in Algorithm \ref{algorithm}.

\textbf{\textit{Source knowledge calibration.} } 
To address the limitation 1, instead of directly inheriting all source knowledge, we design a transferability-controlled knowledge distillation loss $\mathcal{L}_{kd}$, which used $UTR_D$ to control the knowledge distillation by quantifying different channels' transferability and assigning more transferable channels larger weights. On the one hand, it prompts the target model to learn transferable source knowledge and discard less-transferable ones.
On the other hand, it constrains the updated target model by unceasingly distilling the transferable source knowledge along the whole training process, rather than at the beginning only. 
% On the other hand, it allows the target model to unceasingly absorb the knowledge during training rather than only at the beginning, which helps to reserve the transferable source knowledge.

%The quantified knowledge distillation promotes inheriting transferable features from $h_s$ to the target encoder $h_t$.
%In this paper, we focus on the knowledge contained in the source encoder $h_s$. 
%We first propose the Domain Distributional Representation Uncertainty ($UTR_D$) module to estimate the transferability of $h_s$ by producing a $d-dimension$ vector $$UTR_D$(h_s)$, each of its dimensions representing how transferable that dimension is. Then, 

%The source representations learned from source data may be deficient for target domain data. We still need to learn target-specific representation based on the inherited representations. To this end, we propose to infer the category of the target data $\hat{y}$ based on the inherited knowledge. %Given $x_t$, the inferred category is $\hat{y}=\arg\max_y P(y|x_t)$, where $p(y|x_t)=\sigma( g_{t;\theta_g} \circ h_{t;\theta_h}(x_t))$, $\sigma$ is the softmax function. 
%However, due to the cross-domain discrepancies, there is a risk in adapting the model using the inferred category information.

%After inheriting the knowledge from the source model, we obtained a target model $\theta_t=g_t \circ h_t$. \chensays{why is it useful here?}

\textbf{\textit{Target semantics calibration.} } The less-transferable knowledge is prone to lead to incorrect semantics (labels) inferred by the model.
Considering that it is fundamental in SFUDA to update the target model based on the inferred semantics of target instances, calibrating the incorrect semantics is essential to learn a discriminative target model. Specifically, after inferring target semantics (using the source model at the beginning of training, and the target model later), we use the $UTR_I$ to select instances whose inferred semantics are unreliable. Then, a semantic calibration loss $L_{sc}$ is designed to calibrate their model predictions. Specifically, on the one hand, as the semantics inferred by the feature of less-transferable channels tend to be wrong, we proposed to use to ``forget'' the current semantics by minimizing a negative cross-entropy  $L_c$. It implicitly guides the model to re-initialize the parameters representing the less-transferable knowledge. On the other hand, we minimize the entropy of the prediction probability distribution of these instances to force their predictions close to a new and appropriate class category. This procedure discovers the new and proper semantics of these instances.

\textbf{\textit{Adaptation.}} With the above two steps, the target model ``safely'' integrates the source knowledge and target semantics. The adaptation step finally refines the target model using inferred semantics from the transferable knowledge, therefore summarizing a better-performing discriminative model.

\vspace{-5mm}
\subsection{Source Knowledge Calibration and Distillation}
\vspace{-2mm}
Not all source knowledge is  transferable and discriminative to the target domain.
Directly transferring all source knowledge to the target model without dealing with the less-transferable parts of it is detrimental to the adaptation of the target domain.
To this end, we instruct the target model to selectively learn the features of transferable channels of the source encoder, therefore, to inherit transferable knowledge from the source encoder.
Given the source encoder $h_{s}$, the $UTR_D(h_{s})$ is calculated following the Equation \eqref{eqa:UTR_D} to estimate the transferability of each channel in $h_{s}$, where a lower $UTR_D$ value suggests stronger transferability. 
Then, the target model learn the source knowledge based on the identified transferability. We proposed a novel transferability-controlled knowledge distillation loss as the objective: 
\begin{equation}
L_{kd}=\mathbb{E}_{x_t\sim \mathcal{X}_t}  [ \Vert Q(UTR_D(h_{s})) \odot [h_{s}(x_t)- h_{t}(x_t)]\Vert_2],
\end{equation}
where $Q(x)=sigmoid(-x)$ is a monotone minus function, $\odot$ is the Hadamard product. The $Q(UTR_D(h_{s}))$ weights the mean squared error term $\Vert h_{s}(x_t)- h_{t}(x_t)\Vert_2$ to distill knowledge within $h_s$ to $h_t$, aiming to assign large weights to features with low $UTR_D$ while small ones to those with high $UTR_D$, guiding the target model to learn more transferable knowledge from the source model and discard less-transferable ones in a well-controlled manner.

\vspace{-5mm}
\subsection{Target Semantics Calibration} 
\vspace{-2mm}
Refining a target model based on the inferred semantics (labels) of target instances is a fundamental and important step for the adaptation in SFUDA. Due to the less-transferable source knowledge, the predicted target semantics may be incorrect, which greatly hinders the adaptation to the target domain. To this end, we design the target semantics calibration module to calibrate the target semantics.

First, the inferred semantics of a target instance $x_t$ is $\hat{y}=\arg \max p(x_t)$ with probability $p^{\hat{y}}(x_t)$, where  $p(x_t)=\sigma(\theta_s(x_t)/\theta_t(x_t))$ is the source/target model predicted probability distribution, $\sigma(.)$ is the softmax function. Note that we use the source model to infer target semantics at the first epoch, and turn to use the target model later since the target model will be more discriminative to the target domain after adaptation. 

 %The expected knowledge should be able to assign clear and correct semantic information to target data. 
%Let $p(x_t)$ denote the model prediction distribution, i.e. $p(x_t)=\sigma( g_{t;\theta_g} \circ h_{t;\theta_h}(x_t))$, where $\sigma$ is the softmax function.
Second, we leverage $UTR_I$ to detect risk target instances whose semantic is prone to be incorrectly inferred that satisfies  $\{x_t:UTR_I(x_t)>\tau\}$ as $\mathcal{X}_{t;risk}$, where $\tau$ denotes the threshold. Following the first step, the feature encoder that calculates $UTR_I(x_t)$ (Equation \ref{eqa:UTR_I}) changes from $h_s$ to $h_t$ after the first epoch.

Third, based on the detected instances $\mathcal{X}_{t;risk}$, we propose a semantics calibrated loss $L_{sc}$ to calibrate the semantics of these instances. 
Since their  semantics is prone to be inaccurate, we train the target model to firstly forget  these  semantics by minimizing the negative cross-entropy loss. The forget objective $L_f$ is represented  as follows:
{\setlength\abovedisplayskip{1pt}
\setlength\belowdisplayskip{1pt}
\begin{equation}
%\begin{aligned}
    L_{f} =\mathbb{E}_{x_t \sim \mathcal{X}_{t;risk} } -CE(x_t,\hat{y}).
        %\\&+  \sum_{k=1}^K  p^k log p^k;
%\end{aligned}
\end{equation}}As illustrated in Fig. \ref{fig:method} (b.3), optimizing this term 
decreases the prediction probability to the misclassified category $\hat{y}$.

On the other hand, we guide the target model to discover the true semantic by the following discover objective $L_d$: 
{\setlength\abovedisplayskip{1pt}
\setlength\belowdisplayskip{1pt}
\begin{equation}
%\begin{aligned}
    L_{d} = -\mathbb{E}_{x_t \sim \mathcal{X}_{t} }\sum_{k=1}^K p(x_t) log p(x_t),%\sum_{k=1}^K \sigma(\theta_t (x_t)) log \sigma(\theta_t (x_t)).
        %\\&+  \sum_{k=1}^K  p^k log p^k;
%\end{aligned}
\end{equation}}where $p(x_t)=\sigma(\theta_t(x_t))$ is the target model predicted probability distribution.
$L_d$ aims to minimize the entropy of the $p(x_t)$, thus guiding the model to assign the prediction to an appropriate  class.
Note that, instead of only minimizing the entropy on $\mathcal{X}_{t;risk}$, we calculate $L_d$ on all target instances $\mathcal{X}_{t}$. In this way, the semantic information of instances with low $UTR_I$, where the model tends to make the right predictions, is also introduced to help the semantic discovery of instances in $\mathcal{X}_{t;risk}$.

Such a "forget-discover" process  implicitly guides the model to free itself from the shackles of less-transferable knowledge and  facilitates the discovery of the true semantics of the target data, and the semantic calibration loss can be denoted as:
{\setlength\abovedisplayskip{1pt}
\setlength\belowdisplayskip{1pt}
\begin{equation}
    L_{sc}= \gamma L_f+L_d,
\end{equation}}
where $\gamma$ is the scale coefficient of the $L_f$.
\begin{algorithm} 
\caption{Calibrated Adaptation Framework}
\label{algorithm}
%\hspace*{0.02in} 
\begin{algorithmic} 
\REQUIRE Source model parameterized by $\theta_s=g_{s} \circ h_{s}$, target model parameterized $\theta_t=g_{t} \circ h_{t}$, unlabeled target instances $D_t$ 

\REQUIRE hyperparameter $\tau$, $\lambda$, $\gamma$

Calculate $UTR_D(h_s)$
\WHILE{i $<$ max epoch}
\STATE In the $i^{th}$ epoch
\STATE \hspace{0.5cm} Sample batch $T$ from $D_t$ \\
\STATE \hspace{0.5cm} Calculate the $L_{kd}$\\
\STATE \hspace{0.5cm} Infer target semantics\\
\STATE \hspace{0.5cm} Calculate $UTR_I(x_t)$, select $\mathcal{X}_{t;risk}$ with $\tau$\\
\STATE \hspace{0.5cm} Calculate $\gamma L_f+L_d$\\
\STATE \hspace{0.5cm} Train the target model by optimizing $\lambda L_{kd}+ \gamma L_f+L_d$\\
\STATE In the $i+1^{th}$ epoch
\STATE \hspace{0.5cm} Sample batch $T$ from $D_t$ \\
\STATE \hspace{0.5cm} Infer target semantics\\
\STATE \hspace{0.5cm} Calculate $L_a$\\
\STATE \hspace{0.5cm} Train the target model by optimizing $L_a$\\
\STATE  $i=i+2$
%\STATE Split dataset into four parts according to $DRU_(x_t),\tau_{hi},\tau_{li},\tau_p$.
%\STATE Inherit: distill transferable representations from $h_{s}$ to $h_{t;\theta_h}$ via $DRU_(z)$.
%\STATE Calibrate inherited knowledge by Equation \ref{}.
%\STATE Discover new knowledge by Equation \ref{}.

%\STATE Self-discover based on the support set
\ENDWHILE 
\end{algorithmic}  
\end{algorithm}

\vspace{-4mm}
\subsection{Adaptation} 
\vspace{-2mm}
With the above two steps to calibrate the source knowledge and target semantics, the target model then can safely adapt to the target model. 
In the adaptation step, we re-infer the target semantics by the model and use it to refine the target model, ultimately adapting the model to the target domain.

%The target semantics can be easily generated by most existing SFUDA methods such as \cite{liang2020we, yang2021generalized, yang2021exploiting,li2021imbalanced,xia2021adaptive}.
In this step, we adopt the pseudo-label strategy in \cite{liang2020we} to re-infer the semantic $\hat{y}$ of the target instance $x_t$ consider its simplicity and effectiveness.
Given  $x_t$  and  $\hat{y}$, we optimize the model with the cross-entropy loss and the objective of the adapt step can be formulated as:
{\setlength\abovedisplayskip{1pt}
\setlength\belowdisplayskip{1pt}
\begin{equation}
%\begin{aligned}
   L_{a}= \mathbb{E}_{x_t \sim \mathcal{X}_{t} } CE(x_t,\hat{y}).
%\end{aligned}
\end{equation}}
\vspace{-10mm}
\subsection{Training Steps}
\vspace{-2mm}
In this subsection, we summarize the training steps of CAF framework. The two calibration steps are separate with the adapt step. Specifically, in the $i^{th}$ epoch, perform two calibration steps to calibrate transferable source knowledge and target semantics  by:
{\setlength\abovedisplayskip{1pt}
\setlength\belowdisplayskip{1pt}
\begin{equation}
    \min_{\theta_t} \lambda L_{kd}+ \gamma L_f+L_d,
\end{equation}}
where $\lambda$ and $\gamma$ is the scale coefficient.

And in the $i+1^{th}$ epoch, conduct the adaption step to adapt the target model to the target domain:
{\setlength\abovedisplayskip{1pt}
\setlength\belowdisplayskip{1pt}
\begin{equation}
    \min_{\theta_t} \lambda L_{a}.
\end{equation}}

\vspace{-2mm}
\section{Results}
\vspace{-2mm}
\subsection{Datasets}
\vspace{-1mm}
We evaluate our SFUDA method using the following three benchmarks: Office-31 \cite{saenko2010adapting}, the Office-Home\cite{venkateswara2017deep} and the VisDA\cite{peng2017visda}. Office-31\cite{saenko2010adapting} contains 4,652 images in 31 categories from three domains: Amazon (A), Webcam (W) and DSLR (D). 
Office-Home\cite{venkateswara2017deep} consists of four domains, i.e., Artistic images (Ar), Clip Art (Cl), Product images (Pr), and Real-World images (Rw), with 65 classes and a total of 15,500 images. 
VisDA\cite{peng2017visda} is a more challenging dataset, whose source domain contains 152k synthetic images generated by rendering 3D models while the target domain has 55k real object images sampled from Microsoft COCO \cite{lin2014microsoft}.
\vspace{-3mm}
\vspace{-2mm}
\subsection{Implementations}
\vspace{-2mm}
For fair comparisons with existing methods, we adopt the backbone of ResNet-50 \cite{he2016deep} for Office-31 and Office-Home and ResNet-101 for VisDA. Following the setups in \cite{liang2020we,yang2021exploiting}, along with the backbones, we used a fully-connected (fc) layer with the output channels of $256$ as the encoder. A fc layer with the weight normalization as the classifier. The source model is trained following the same strategy with \cite{liang2020we,yang2021exploiting}.
\textbf{The pre-trained source model is used to adapt to the target domain but without using any labeled source data.}
In the optimization, we adopt SGD with momentum 0.9 and batch size of 64 on all datasets. For the Office-31 and office-Home datasets, the learning rates used to train the ResNet-50 backbone and the newly added layers are 1e-3 and 1e-2 respectively. The learning rate is 1e-4 for VisDA. We trained 40, 60 and 50 epochs for Office-31, Office-Home and VisDA respectively. Note that the mixup\cite{zhang2017mixup} data augmentation is used in the adaptation step. The threshold of $UTR_I$, i.e. $\tau$, is set to be 3. The weight  $\lambda$ of the transferability-controlled knowledge distillation loss is set to 10 at the beginning. As the training procedure progresses, the model is gradually adapted to the target domain, requiring less source knowledge. Therefore, after 10 epochs, we decrease $\lambda$ to zero. The weight  $\gamma$ of the "forget" loss is set to 0.9. 
For the uncertainty measurement (Equation \ref{eqa:mUTR_D}), $T$ is set to $2$, and $r_t$ is randomly sampled from the uniform distribution $U(-0.05,0.05)$.

\begin{table}
\caption{Classification accuracies (\%) on  Office-31 dataset.}
\centering
\resizebox{\linewidth}{!}{
\begin{tabular}{l|c|c|c|c|c|c|c}
\hline

Method&A$\rightarrow$D &A$\rightarrow$W &D$\rightarrow$A &D$\rightarrow$W &W$\rightarrow$A &W$\rightarrow$D&Avg.\\
\hline
Source-model&80.4 &76.5&60.2 &95.6 &63.4 &98.6 &79.1\\
SoFA\cite{yeh2021sofa}  &73.9 &71.7 &53.7& 96.7 &54.6 &98.2& 74.8\\
SFDA\cite{kim2021domain}   &92.2 &91.1 &71.0 &98.2 &71.2 &99.5& 87.2\\

SHOT\cite{liang2020we}    &94.0 &90.1 &74.7 &98.4 &74.3 &99.9& 88.6\\
3C-GAN\cite{li2020model}  &92.7 &93.7 &75.3 &98.5 & 77.8& 99.8 &89.6\\
BAIT\cite{yang2020unsupervised}   &92.0 &\textbf{94.6} &74.6 &98.1  & 75.2 &\textbf{100.0} &89.1\\
NRC\cite{yang2021exploiting} & \textbf{96.0} &90.8& 75.3 &99.0 &75.0 &\textbf{100.0} & 89.4\\
HCL\cite{huang2021model} & 94.7& 92.5 &75.9 &98.2& 77.7& 100&89.8\\
AAA\cite{li2021divergence}  &95.6 &94.2 &75.6& 98.1 &76.0 &99.8& 89.9\\
A2Net \cite{xia2021adaptive}  &94.5 &94.0 &\textbf{76.7}& \textbf{99.2} & 76.1 &\textbf{100.0}& 90.1\\
DIPE\cite{wang2022exploring}   &96.6 &93.1 &75.5 &98.4& 77.2 &99.6 &90.1\\
Ours  &95.0 &93.5 &76.3 & 99.1& \textbf{78.4}& \textbf{100.0} &\textbf{90.3}\\

\hline
\end{tabular}
}

\label{tab_31}
\end{table}

\vspace{-4mm}
\subsection{Comparison with State-of-the-Art Methods}
\vspace{-2mm}
We report the results on Office-31, Office-Home, and VisDA, in Tables  \ref{tab_31}, \ref{tab_home}, and \ref{tab_visda}, respectively. 

On Office-31 tasks, in terms of the average accuracy of 6 transfer tasks, our method outperforms the state-of-the-art work A2Net\cite{xia2021adaptive} and DIPE\cite{wang2022exploring} by 0.2\%, improving from 90.1\% to 90.3\%. We also achieve the state-of-the-art results on W$\rightarrow$A and W$\rightarrow$D. For other transfer directions of Office-31, we achieved very competitive results. We hypothesize the reason of the results is that our method brings transferability risk quantification to SFUDA and integrates the “safe-to-transfer” source knowledge to the target domain for better adaptation. We also argue that our method is more useful and brings more improvements for challenging adaptation tasks, where the cross-domain transfer risk is high. \textit{Instead,} the Office-31 transfer tasks are easy and bring less risks (considering that the average accuracy of the source-only model is 79.1\%), so our method improvement is competitive and not that significant.

As expected, on the more challenging Office-Home tasks and VisDA tasks (the mean accuracy of the source models are 60.0\% and 48.0\%), our method brings larger improvement. In the Office-Home tasks, we achieve the state-of-the-art performance on 8 of 12 tasks, and also outperform the prior work in terms of the average accuracy of 0.6\%, improving from 72.6\% (by A2Net\cite{xia2021adaptive}) to 73.2\%. Particularly, we have achieved significant improvements in two difficult tasks Ar$\rightarrow$Cl and  Re$\rightarrow$Cl and outperform the second best one by 1.2\% and 0.7\%, respectively. On the VisDA tasks, our method outperforms others among 10 out of 12 tasks and surpasses the SOTA method NRC by a large margin (2.4\%) in the per-class accuracy.

Compared with the most related work DIPE\cite{wang2022exploring}, our method also obtains  performance improvement in all three benchmarks, including 0.2\% in Office-31, 0.7\% in Office-Home and 4.2\% in VisDA. The reported results clearly demonstrate the efficacy of our method.

\begin{table}\scriptsize
\vspace{-2mm}
\caption{Classification accuracies (\%) on  Office-Home dataset (ResNet-50). AC denote the task Ar$\rightarrow$CL.}
\vspace{-2mm}
\setlength\tabcolsep{0.55pt}
\centering
\resizebox{\linewidth}{!}{
%\fontsize{7.7}{8.5}\selectfont
\begin{tabular}{l|c|c|c|c|c|c|c|c|c|c|c|c|c}
\hline

%Method&SF&Ar$\rightarrow$CL &Ar$\rightarrow$Pr &Ar$\rightarrow$Re &Cl$\rightarrow$Ar &Cl$\rightarrow$Pr &Cl$\rightarrow$Re &Pr$\rightarrow$Ar &Pr$\rightarrow$Cl &Pr$\rightarrow$Re &Re$\rightarrow$Ar &Re$\rightarrow$CL &Re$\rightarrow$Pr&Avg.\\
Method&AC &AP &AR &CA &CP &CR &PA &PC &PR &RA &RC &RP&Avg.\\
\hline
Source-model&44.8 &67.4 &75.1 &52.3 &63.4 &63.7 &53.6 &39.5 &72.7 &64.1 &45.2 &77.6 &60.0\\
SHOT\cite{liang2020we}   &57.1 &78.1 &81.5 &68.0 &78.2 &78.1& 67.4 &54.9 &82.2 &73.3 &58.8& 84.3 & 71.8\\
SFDA\cite{kim2021domain}  &48.4 &73.4 &76.9& 64.3& 69.8 &71.7 &62.7 &45.3& 76.6 &69.8 &50.5 &79.0& 65.7\\
SoFA\cite{yeh2021sofa}  &- &74.1&77.6 &- &71.8 &75.1 &- &- &- &- &- &- &-\\
BAIT\cite{yang2020unsupervised}    &57.4  &77.5  &82.4  &68.0 & 77.2 & 75.1 & 67.1 & 55.5 & 81.9  &73.9  &59.5  &84.2  &71.6\\
PS \cite{du2021generation}   &57.8&77.3 &81.2& 68.4 &76.9 &78.1 &67.8 &\textbf{57.3}& 82.1 &75.2& 59.1&83.4& 72.1\\
AAA\cite{li2021divergence} &56.7 &78.3& 82.1& 66.4& 78.5& 79.4& 67.6& 53.5& 81.6 &74.5& 58.4 &84.1 &71.8\\
NRC\cite{yang2021exploiting} & 57.7& 80.3 &82.0 &\textbf{68.1}& \textbf{79.8} &78.6 &65.3 &56.4 &\textbf{83.0}& 71.0 &58.6& 85.6 &72.2\\
DIPE  \cite{wang2022exploring} &56.5& 79.2& 80.7& 70.1 &79.8 &78.8& 67.9& 55.1 &83.5 &74.1 &59.3& 84.8 &72.5\\
A2Net \cite{xia2021adaptive}  &58.4 &79.0& 82.4 &67.5 &79.3& 78.9 &68.0 &56.2& 82.9 &\textbf{74.1} &60.5 &85.0 &72.6\\
Ours  &\textbf{59.8} &\textbf{81.2}&\textbf{83.2}  &67.2 &79.2 &\textbf{80.1} &\textbf{ 68.4}&56.4&\textbf{83.0}  &73.7&\textbf{61.2} &\textbf{85.9}&\textbf{73.2}\\

\hline
\end{tabular}}

\label{tab_home}
\end{table}

\begin{table}\scriptsize
\vspace{-4mm}
\caption{Classification accuracies (\%) on  VisDA-C dataset  (ResNet-101), $Per_c$ denotes the per-class accuracy.}
\vspace{-2mm}
\setlength\tabcolsep{0.55pt}
\centering
\resizebox{\linewidth}{!}{
\begin{tabular}{l|c|c|c|c|c|c|c|c|c|c|c|c|c}
\hline
Method&plane &bcycl &bus &car &horse &knife &mcycl &person &plant &sktbrd &train &truck&$Per_c$\\
\hline
Source-model&64.6 &28.9 &47.2 &63.5 &67.2 &12.4 &82.5 &23.5 &61.7 &31.4 &82.1 &11.1 &48.0\\
3C-GAN\cite{li2020model}    &94.8  &73.4  &68.8  &74.8  &93.1 & 95.4  &88.6 & \textbf{84.7} & 89.1 & 84.7 & 83.5  &48.1  &81.6\\
SHOT\cite{liang2020we}  &94.3 &88.5 &80.1 &57.3& 93.1& 94.9& 80.7& 80.3 &91.5 &89.1 &86.3 &58.2& 82.9\\
SFDA\cite{kim2021domain}  &86.9 &81.7 &84.6 &63.9 &93.1& 91.4 &86.6& 71.9& 84.5& 58.2 &74.5& 42.7 &76.7\\
BAIT\cite{yang2020unsupervised}   &93.7 &83.2& 84.5& 65.0 &92.9 &95.4 &88.1 &80.8 &90.0 &89.0 &84.0 &45.3&82.7\\
DIPE\cite{wang2022exploring} & 95.2& 87.6 &78.8 &55.9& 93.9 &95.0 &84.1 &81.7 &92.1 &88.9 &85.4 &58.0 &83.1\\
HCL\cite{huang2021model} &93.3& 85.4& 80.7 &68.5& 91.0& 88.1 &86.0& 78.6& 86.6& 88.8& 80.0& \textbf{74.7}&83.5\\
PS\cite{du2021generation}  &95.3 &86.2 &82.3 &61.6 &93.3 &95.7& 86.7& 80.4& 91.6& 90.9& 86.0 &59.5& 84.1\\
AAA\cite{li2021divergence} &94.4 &85.9 &74.9 &60.2 &96.0& 93.5& 87.8& 80.8& 90.2& 92.0 &86.6 &68.3 &84.2\\

A2Net \cite{xia2021adaptive}  &96.1 &88.3& 85.5 &74.1 &97.1 &95.4& 89.5 &79.4 &95.4 &92.9& 89.1& 42.6& 85.4\\

NRC\cite{yang2021exploiting} &96.8& 91.3& 82.4 &62.4 &96.2 &95.9 &86.1 &80.6& 94.8& \textbf{94.1} &90.4 &59.7 &85.9\\
Ours  &\textbf{98.0} &\textbf{92.9} &\textbf{88.3} &\textbf{78.0}&\textbf{97.8} &\textbf{97.7}& \textbf{91.1 }& \textbf{84.7} &\textbf{95.5}  &91.4& \textbf{91.2}& 41.1&\textbf{87.3}\\

\hline
\end{tabular}}

\label{tab_visda}
\vspace{-4mm}
\end{table}

\begin{figure}[t]
  \centering
 \vspace{-2mm}
    \includegraphics[width=1.0\linewidth]{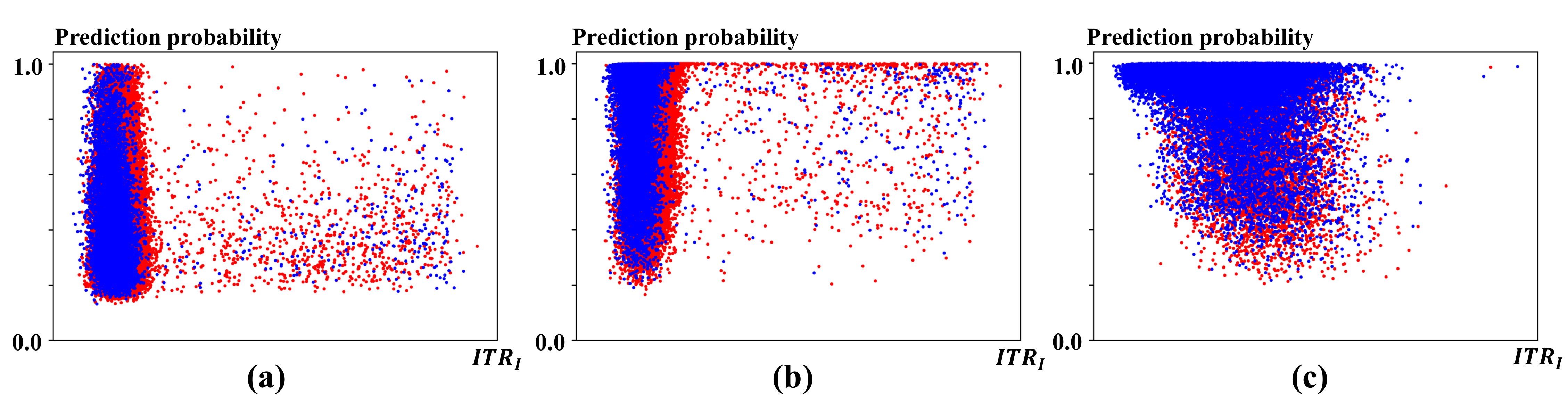}
\vspace{-6mm}
   \caption{The visualization of the prediction probability, prediction accuracy  of different samples and their $UTR_I$  in VISDA by (a): source model (b) CAF without $L_f$, and (c) CAF.  Blue point represent samples that the model predicted correctly, and red indicates that the prediction is wrong. The vertical axis represents the prediction probability of the sample, and the horizontal axis represents $UTR_I$.}
   \label{fig:visual}
\end{figure}
\begin{figure}

 \centering
  
\includegraphics[width=0.6\linewidth]{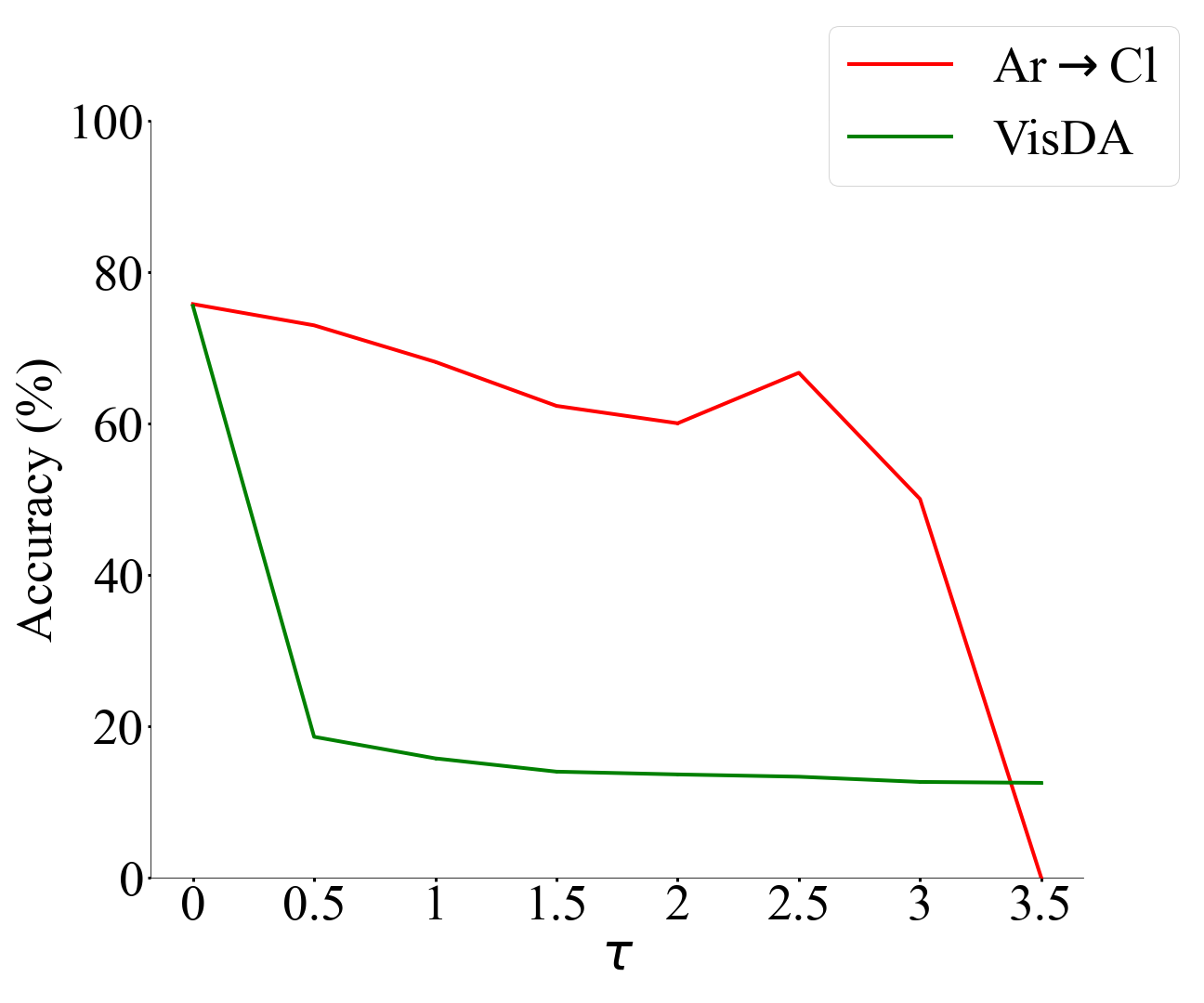}
\vspace{-1mm}
\caption{The Accuracy-$UTR_I$ curve of source model to target samples on Ar$\rightarrow$Cl and VisDA tasks. The horizontal axis is the threshold of $UTR_T$, denotes samples that satisfied  $UTR_I(x_t)>\tau$. The vertical axis represents the predicted semantic accuracy of the model for these samples. For a better illustration, we select samples to which the max prediction probability of the source model are larger than 0.5.}
\label{fig:acc_UD_UTR}
\vspace{-5mm}
\end{figure}

\begin{figure}
  \centering

\includegraphics[width=0.3\linewidth]{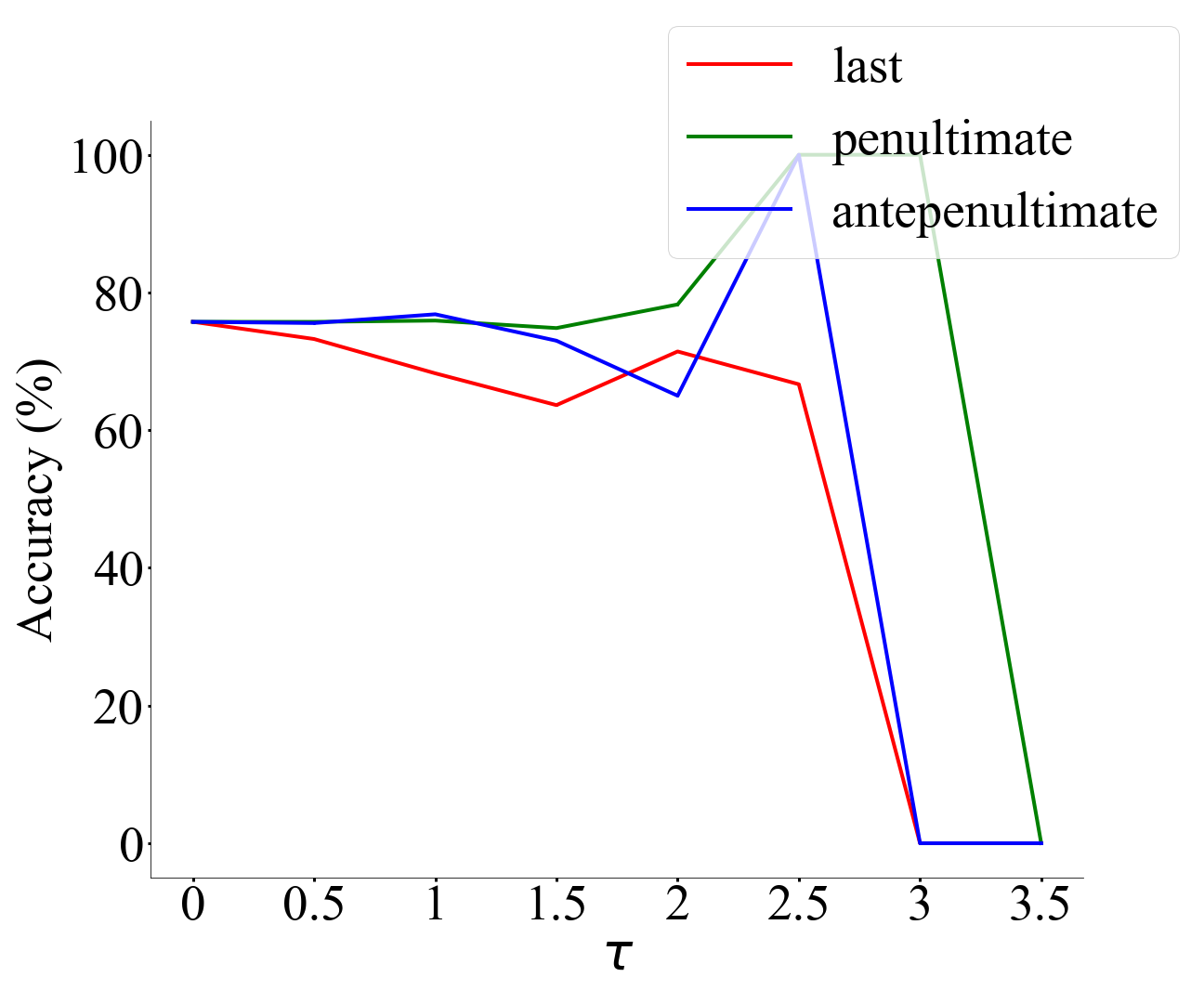}

\includegraphics[width=0.3\linewidth]{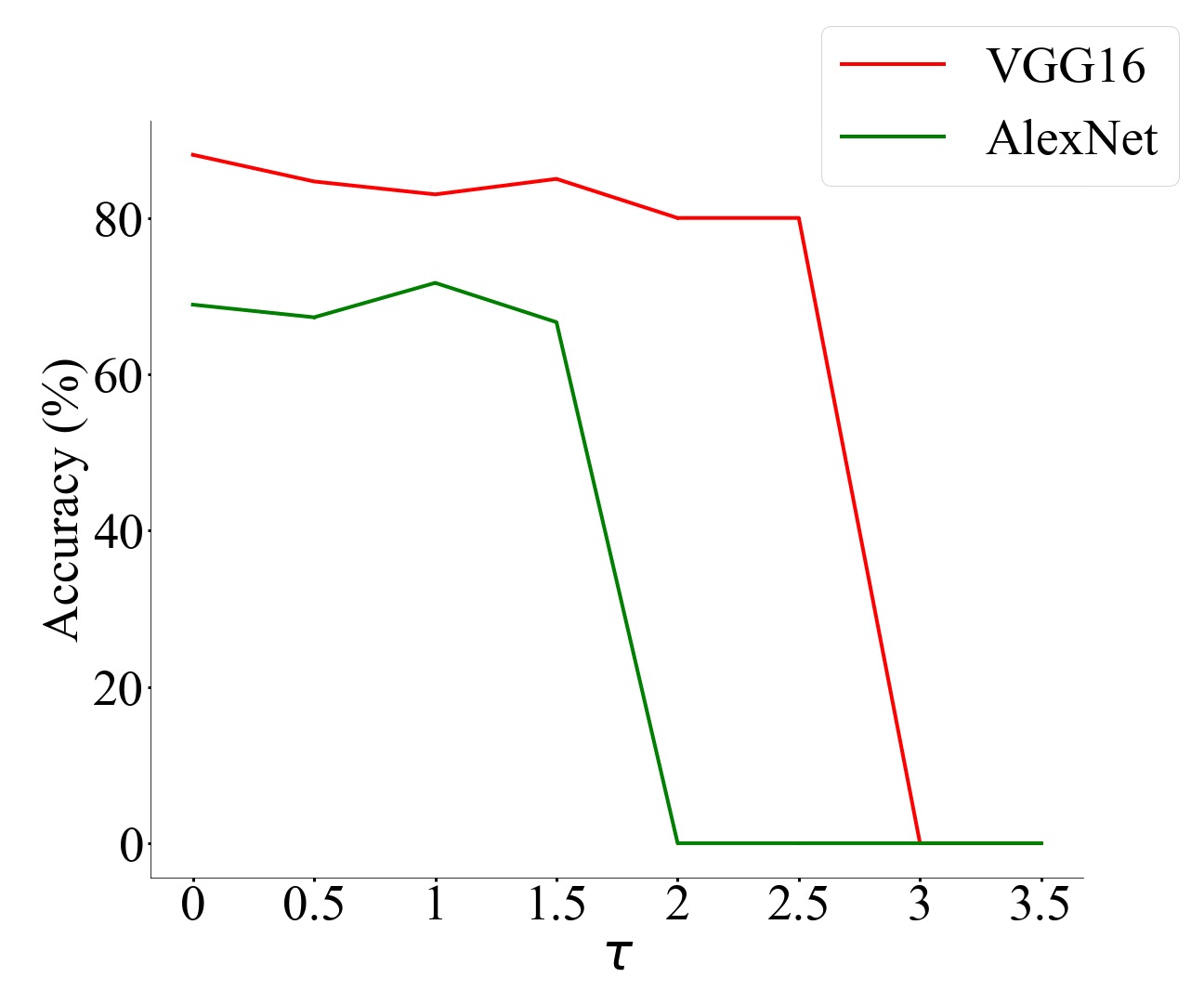}

\includegraphics[width=0.3\linewidth]{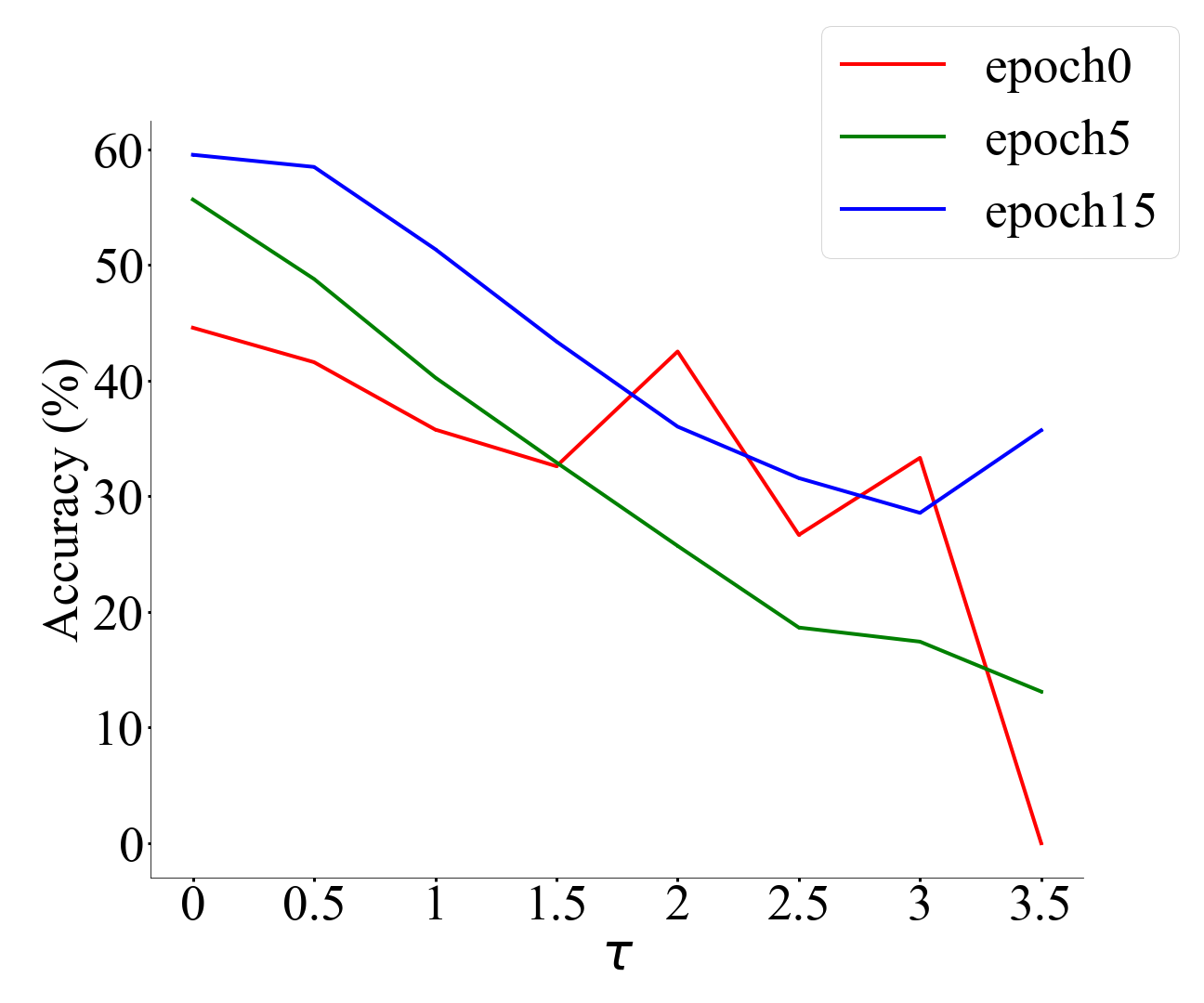}

\vspace{-1mm}
\caption{The Accuracy-$UTR_I$ curves in Ar$\rightarrow$Cl where the $UTR_I$ is calculated on (a) different layers ( the last, penultimate, and antepenultimate bottlenecks of Resnet-50), (b) other model structures (VGG16 and AlexNet) and (c) the target model.  For a better illustration, we report the prediction accuracy of the source model for samples where the max prediction probability are larger than 0.5, 0.9 for (a) and (b), respectively.}
\label{fig:acc_other_UTR}

\end{figure}

%%%%%%%%%%%%%%%%%%

% \begin{figure*}
% %   \centering
%   %\fbox{\rule{0pt}{2in} \rule{0.9\linewidth}{0pt}}
% \begin{subfigure}[b]{0.24\textwidth}
% % \vspace{-5mm}
% \centering
% \includegraphics[width=0.24\textwidth]{figs/sourceac.png}
% \vspace{-5mm}
% % \caption{Source}
% \end{subfigure}%
% \vspace{-5mm}
% \hspace{-6mm}
% \begin{subfigure}[b]{0.24\textwidth}
% % \vspace{-5mm}
% \centering
% \includegraphics[width=0.24\textwidth]{figs/shotac.png}
% \vspace{-5mm}
% \end{subfigure}%
% \vspace{-5mm}
% \hspace{-6mm}
% \begin{subfigure}[b]{0.24\textwidth}
% % \vspace{-5mm}
% \centering
% \includegraphics[width=0.24\textwidth]{figs/nrcac.png}
% \vspace{-5mm}
% \end{subfigure}%
% \vspace{-5mm}
% \hspace{-6mm}
% \begin{subfigure}[b]{0.24\textwidth}
% % \vspace{-5mm}
% \centering
% \includegraphics[width=0.24\textwidth]{figs/oursac.png}
% \vspace{-5mm}
% \end{subfigure}%
% \vspace{-5mm}

% \vspace{-3mm}
% \caption{The t-SNE  visualizations of different methods on the target domain Ar of the Office-Home task Cl$\rightarrow$Ar, including: the source model, Shot, NRC and Ours. For a better illustration, we choose features in the first 6 classes, and different color denotes different class. Best viewed in colors.}
% \label{fig:tSNE}

% \end{figure*}

%%%%%%%%%%%%%%%%%%

%\begin{figure}[t]
%  \centering
%  %\fbox{\rule{0pt}{2in} \rule{0.9\linewidth}{0pt}}
 
%    \includegraphics[width=0.9\linewidth]{figs/visda.eps}

%   \caption{ Performance on different parts \chensays{remember to adjust the font size. otherwise, you will always spend more time in the final day to re-draw the figures.}}
%   \label{fig:acc_curve}
%\end{figure}
\vspace{-4mm}
\section{Analysis}

%\chensays{In the analysis section, we need basically two sub-section: quantitative analysis and qualitative analysis. The aim is to show that 1) how good is our method; 2) why our method is good.}

%\chensays{Finally, any study has limitations and every paper has some insights to shout aloud to the community. These things, e.g. summary and conclusions should be in the discussion section.}

\vspace{-2mm}
\subsection{Ablation Study}
\vspace{-1mm}
\noindent \textbf{Ablation study on the Source Knowledge Calibration.} 
We designed the transferability-controlled knowledge distillation loss $L_{kd}$ to distill transferable source knowledge. To prove its effectiveness, the ablation results of this module are reported in the first three rows of Table \ref{tab_abl_module}. It can be seen that the v2 ($L_a + L_{kd}$) outperforms v1 ($L_a$) by 2.1\% in Ar$\rightarrow$Cl and 1.7\% in Ar$\rightarrow$Re respectively. However, it induces negative effects on Ar$\rightarrow$Pr. We hypothesize the reasons to be that the transferable knowledge to the target domain may not discriminate the target samples and may infer wrong semantics in the Ar$\rightarrow$Pr. To prove our hypothesis, by comparing results of v4 and ours in Table \ref{tab_abl_module}, it proves that by adding two semantics calibration losses $L_f$ and $L_d$, the $L_{kd}$ is more effective and brings significant improvements on all transfer directions Ar$\rightarrow$Cl, Ar$\rightarrow$Re and Ar$\rightarrow$Pr (see more details in the following two comparisons: [v2 $\leftrightarrow$ v3] and [v4 $\leftrightarrow$ Ours]). It may prove that calibrating the target semantics helps calibrate transferable source knowledge.

\begin{table}[t]\scriptsize
\vspace{-4mm}
\caption{Ablation study on three Office-Home task. }
\vspace{-2mm}
\centering
\resizebox{\linewidth}{!}{
\begin{tabular}{l|c|c|c|cc}
\hline
Method&Module &Ar$\rightarrow$Cl &Ar$\rightarrow$Pr &Ar$\rightarrow$Re\\
\hline
source&- &44.6& 67.3 &74.0\\
v1 &$L_{a}$&50.4 & 73.3  & 75.6\\
v2 &$L_{a}+L_{kd}$&52.5 & 72.4 &77.3 \\
v3 &$L_{a}+L_{d}$& 54.6& 77.1 &80.6\\
v4 &$L_{a}+L_{f}+L_{d}$ &57.1&77.2&81.1\\
v5 & $L_{a}+L_{d}+L_{kd}  $ &58.5 & 79.2 & 82.0 & \\
v6 & $L_{a}+L_{f}+L_{kd}  $ &46.6 &70.1 &74.3& \\

Ours &$L_{a}+L_{f}+L_{d}+L_{kd}$ & 59.8 & 81.2 & 83.2 & \\
\hline
Ours-Merge&$L_{a}+L_{f}+L_{d}+L_{kd}$&57.7&79.3&82.1\\
\hline
Ours-Online&$L_{a}+L_{f}+L_{d}+L_{kd}$&59.2&81.2&83.1\\
\hline
Ours-Ensemble &$L_{a}+L_{f}+L_{d}+L_{kd}$&59.1&80.7&82.9\\
Ours-MC dropout&$L_{a}+L_{f}+L_{d}+L_{kd}$&59.3&80.5&82.7\\
%w/o L_{kd} &56.1&77.2&81.1\\
%w/o L_{cl}&56.8&75.5&78.1\\
%w/o L_{dis}&46.6&70.1&74.3\\
\hline
\end{tabular}}

\label{tab_abl_module}

\end{table}

\begin{table}[t]\scriptsize
\vspace{-4mm}
\caption{Hyperparameter(HPR) analysis. The results are shown in form of  Value/Acc(\%).}
\vspace{-2mm}
\centering
\resizebox{\linewidth}{!}{
\begin{tabular}{l|c|c|c|c|c|c}

\hline
HPR &\multicolumn{6}{c}{Ar$\rightarrow$Pr}\\
\hline
%source-only &44.6& 67.3 &74.0\\
%$\lambda_1$ &0.5/78.7 &2/79.5&5/81.2 &10/78.0 &15/77.2&25/75.7  \\
%$\lambda_2$ &0.03/75.8 &0.1/77.7&0.3/80.4 &1/81.2 &3/80.4&10/68.6  \\
$\tau$ &1.5m/76.2&2m/78.6&2.5m/81.2&3m/81.2 &3.5m/80.6 &4m/79.6  \\
$\lambda$ &0.5/78.56 & 2.5/77.91&5/79.08 &7.5/80.1 &10/81.2 &12.5/80.4\\
$\gamma$&0.1/79.3&0.5/80.7&0.9/81.2&1.0/80.8&1.5/79.5&2/79.4 \\
%$\lambda_1$ &0.03/79.5 &0.05/80.8&0.08/81.2 &0.12/81.0 &0.2/78.8&0.3/78.6  \\
%$\lambda_2$ &0.03/79.5 &0.05/80.8&0.08/81.2 &0.12/81.0 &0.2/78.8&0.3/78.6  \\
\hline
\end{tabular}}

\label{tab_hyp}

\end{table}

\noindent \textbf{Ablation study on the Target Semantics Calibration}
We designed a forget loss $L_{f}$ and a discover loss $L_{d}$ to calibrate the target semantics subsequently. 
% Target Semantics Calibration aims to calibrate target semantics, including a forget loss $L_{f}$ to forgot the current prediction and a discover loss $L_{d}$ to self-discover their true category subsequently. 
By comparing the results of v1 and v3 in Table \ref{tab_abl_module}, we may conclude that adding $L_{d}$ in v1 boosts the performances by 4.2\%, 3.8\% and 5.0\% on the three tasks respectively. 

In addition, using the $L_{f}$ only without the $L_{d}$ brings negative effects (see the comparison [v6 $\leftrightarrow$ Ours]). However, from comparisons [v3 $\leftrightarrow$ v4] and [v5 $\leftrightarrow$ Ours] in Table \ref{tab_abl_module}, we verify that $L_{f}$ is only effective when combined with $L_{d}$. 

Moreover, we notice that the forget loss $L_{f}$ is more effective on challenging tasks, e.g. Ar$\rightarrow$Cl. We hypothesize that it is because the predictions on challenging tasks tend to be wrong and therefore forgetting model prediction completely brings more improvements.

To further understand the forget loss $L_{f}$ and prove the previous arguments, we visualize the correct/incorrect case of target prediction and the $UTR_I$ obtained by the source model, our CAF model without the forget loss and our CAF model in Fig. \ref{fig:visual} (a), (b) and (c) respectively. It can be seen in (a) that the source model predicts many wrong semantics in the target domain (the red points) due to the less-transferable knowledge. By observing many red points on the upper right of the Fig. \ref{fig:visual} (b), it seems that we can not calibrate the wrong semantics of samples with high prediction probability without $L_f$, since the model is confident in its inferred semantics and tends to maintain these semantics. Finally, from Fig. \ref{fig:visual} (c), it can be seen that after adding the forget term $L_f$, the model forgets the wrong semantics and finally calibrates their semantics. The above experiments verify the effectiveness of the Target Semantics Calibration. 

\begin{table}
\vspace{-4mm}
\scriptsize
\caption{Improvement to existing SFUDA methods.}
\vspace{-2mm}
\centering

%\fontsize{7.7}{8.5}\selectfont

\begin{tabular}{l|c|c|c}
\hline

%Method&SF&Ar$\rightarrow$CL &Ar$\rightarrow$Pr &Ar$\rightarrow$Re &Cl$\rightarrow$Ar &Cl$\rightarrow$Pr &Cl$\rightarrow$Re &Pr$\rightarrow$Ar &Pr$\rightarrow$Cl &Pr$\rightarrow$Re &Re$\rightarrow$Ar &Re$\rightarrow$CL &Re$\rightarrow$Pr&Avg.\\
Method&Cl$\rightarrow$ Ar &Cl$\rightarrow$Pr &Cl$\rightarrow$Re \\

\hline
Ours&67.2 &79.2 &80.1\\
\hline
SHOT\cite{liang2020we}  &68.0 &78.2 &78.1\\
SHOT+Ours  &68.5 &79.3 &80.3 \\
\hline

NRC\cite{yang2021exploiting} &68.1& 79.8 &78.6\\
NRC+Ours  &68.9 &80.1 &80.2 \\
\hline
\end{tabular}

\label{tab_existing}
\end{table}

\noindent \textbf{Ablation study on merging different steps.}
In our CAF framework, the two calibration steps integrate ``transferable'' source knowledge and the reliable target semantics \textit{first} and \textit{after that} the adaptation step refines the model using the calibrated knowledge and target semantics. Therefore, it is necessary and better to perform the two calibration steps before the adaptation step. To prove the necessary, we conduct extra experiments performing the calibration and adaptation steps simultaneously. The results are denoted as ``Ours-Merge'' in Table \ref{tab_abl_module}. It can be observed that the ``Ours'' result outperforms  the ``Ours-merge'' result by 2.1\%, 1.9\% and 1.1\% in Ar$\rightarrow$Cl, Ar$\rightarrow$Pr and Ar$\rightarrow$Re respectively.

\vspace{-4mm}
\subsection{Hyperparameter Analysis}
\vspace{-2mm}
We analyse the sensitivity of the following hyperparameters: the $UTR_I$ threshold $\tau$ and the weights $\lambda$ and $\gamma$ of the losses $L_{KD}$ and $L_{f}$ respectively. 
The results in Table \ref{tab_hyp} demonstrate that our method is stable to the choices of hyperparameters in a wide range.

\vspace{-4mm}
\subsection{Calibration on other existing methods}
\vspace{-2mm}
Without measuring the transferability, current SFUDA methods \cite{yang2021exploiting,yang2021generalized,xia2021adaptive,yang2020unsupervised,liang2020we} directly perform adaptation but ignore calibration steps in our CAF framework. Our two calibration steps fill in the gap, and are flexible and ``plug-and-play''. Therefore, we add our calibration modules on existing SFUDA works \cite{yang2021exploiting,liang2020we} and report the experimental results in Table \ref{tab_existing}. It can be seen that adding our calibration modules on SHOT/NRC methods improves SHOT/NRC by 0.9/0.8\%, 1.2/0.3\%, and 2.2/2.1\% on Cl$\rightarrow$Ar, Cl$\rightarrow$Pr and Cl$\rightarrow$Re respectively. It proves that our CAF method is ``plug-and-play'' and effective to different SFUDA baselines.

\begin{table*}
\vspace{-3mm}
\caption{Comparison with different transferability measurements on the  Office-Home tasks. $z_{low}$ and $z_{high}$  are features with  low $UTR_D$ and  high $UTR_D$, respectively. The $\uparrow$/$\downarrow$ indicates the larger/smaller the value, the higher the transferability. 
In each task, current transferability measurements are calculated on $Z_{low}$ and $Z_{high}$, respectively.
The more transferable one is bolded.}
\vspace{-3mm}
\centering
\resizebox{\linewidth}{!}{
\begin{tabular}{lcccccccccccccccccccccccc}

\hline
\multirow{3}{*}{Measurement}&\multicolumn{2}{c}{Ar$\rightarrow$Cl}&\multicolumn{2}{c}{Ar$\rightarrow$Pr}&\multicolumn{2}{c}{Ar$\rightarrow$Re}&\multicolumn{2}{c}{Cl$\rightarrow$Ar}&\multicolumn{2}{c}{Cl$\rightarrow$Pr}&\multicolumn{2}{c}{Cl$\rightarrow$Re}&\multicolumn{2}{c}{Pr$\rightarrow$Ar}&\multicolumn{2}{c}{Pr$\rightarrow$Cl}&\multicolumn{2}{c}{Pr$\rightarrow$Re}&\multicolumn{2}{c}{Re$\rightarrow$Ar}&\multicolumn{2}{c}{Re$\rightarrow$Cl}&\multicolumn{2}{c}{Re$\rightarrow$Pr}\\
%\cmidrule(r){2-3}\cmidrule(r){4-5}\cmidrule(r){6-7}\cmidrule(r){8-9}\cmidrule(r){10-11}\cmidrule(r){12-13}\cmidrule(r){14-15}\cmidrule(r){16-17}\cmidrule(r){18-19}
& $Z_{low}$ & $Z_{high}$& $Z_{low}$ & $Z_{high}$& $Z_{low}$ & $Z_{high}$& $Z_{low}$ & $Z_{high}$& $Z_{low}$ & $Z_{high}$& $Z_{low}$ & $Z_{high}$& $Z_{low}$ & $Z_{high}$& $Z_{low}$ & $Z_{high}$& $Z_{low}$ & $Z_{high}$\\
\hline
%source-only &44.6& 67.3 &74.0\\
%$\lambda_1$ &0.5/78.7 &2/79.5&5/81.2 &10/78.0 &15/77.2&25/75.7  \\
%$\lambda_2$ &0.03/75.8 &0.1/77.7&0.3/80.4 &1/81.2 &3/80.4&10/68.6  \\

MMD$\downarrow$&\textbf{0.38}&0.41&0.11&0.11&\textbf{0.50}&0.56&\textbf{0.71}&0.74&\textbf{0.20}&0.21&\textbf{0.17}&0.18&\textbf{0.48}&0.54&\textbf{0.50}&0.57&0.32&\textbf{0.30}&\textbf{0.30}&0.31&0.58&\textbf{0.54}&\textbf{0.19}&0.24 \\
A-Distance$\downarrow$&\textbf{1.47}&1.51&\textbf{1.23}&1.35&\textbf{0.80}&0.86&\textbf{1.40}&1.43&\textbf{1.20}&1.25&\textbf{1.40}&1.44 &\textbf{1.33}&1.45&\textbf{1.47}&1.52&\textbf{0.84}&0.87&\textbf{1.00}&1.07&1.47&\textbf{1.45}&0.85&\textbf{0.88} \\
Corresponding Angle$\uparrow$ &\textbf{-0.14}&-0.15 &\textbf{-0.05}&-0.13&-0.72&\textbf{-0.71}&\textbf{0.97}&0.94 &\textbf{0.98}&0.97&\textbf{0.99}&0.97&\textbf{0.27}&0.09 &\textbf{-0.09}&-0.49&\textbf{0.31}&0.28&\textbf{0.29}&0.21 &-0.12&\textbf{0.40}&\textbf{0.23}&0.08\\
LogME$\uparrow$&\textbf{0.83}&0.82 &\textbf{0.93}&0.92& \textbf{0.89}&0.87&\textbf{0.85}&0.81 &\textbf{0.84}&0.82& \textbf{0.84}&0.81&\textbf{0.83}&0.82 &\textbf{0.81}&0.80& \textbf{0.86}&0.84&\textbf{0.84}&0.83 &\textbf{0.84}&0.82& \textbf{0.92}&0.90\\
LEEP$\uparrow$&\textbf{-3.66}&-3.79 &\textbf{-3.30}&-3.35& \textbf{-3.25}&-3.34&\textbf{-2.81}&-2.98 &\textbf{-2.39}&-2.65& \textbf{-2.38}&-2.54&\textbf{-3.51}&-3.63 &\textbf{-3.75}&-3.85& \textbf{-3.15}&-3.28&\textbf{-3.20}&-3.36& \textbf{-3.51}&-3.61& \textbf{-3.12}&-3.24\\
NCE$\uparrow$&\textbf{-2.05}&-2.17 &\textbf{-1.21}&-1.31& \textbf{-1.12}&-1.51&\textbf{-1.71}&-1.99 &\textbf{-1.43}&-1.58& -1.44&\textbf{-1.43}&\textbf{-1.82}&-2.07 &\textbf{-2.30}&-2.55& \textbf{-1.21}&-1.39&\textbf{-2.43}&-2.54 &\textbf{-2.09}&-2.41& \textbf{-0.93}&-1.05\\
Accuracy(\%)$\uparrow$&\textbf{49.5}&47.1&\textbf{60.3}&58.2&\textbf{62.9}&61.2&\textbf{48.6}&47.2&\textbf{59.5}&57.9&\textbf{61.7}&60.4&\textbf{48.4}&45.3&\textbf{38.7}&33.9&\textbf{68.8}&67.5&\textbf{61.8}&60.2 &\textbf{43.5}&39.6 &\textbf{75.4}&73.1\\
%Accuracy$\uparrow$&\textbf{45.5\%}&42.2\%&\textbf{66.4\%}&64.2\%&\textbf{73.9\%}&70.5\%\\
%$\lambda_1$ &0.03/79.5 &0.05/80.8&0.08/81.2 &0.12/81.0 &0.2/78.8&0.3/78.6  \\
%$\lambda_2$ &0.03/79.5 &0.05/80.8&0.08/81.2 &0.12/81.0 &0.2/78.8&0.3/78.6  \\
\hline
\end{tabular}}

\label{tab_transferability1}

\end{table*}

\begin{table*}

\caption{Comparison with different transferability measurements on the Office-31 and VisDA tasks. $z_{low}$ and $z_{high}$  are features with  low $UTR_D$ and  high $UTR_D$, respectively. The $\uparrow$/$\downarrow$ indicates the larger/smaller the value, the higher the transferability. 
In each task, current transferability measurements are calculated on $Z_{low}$ and $Z_{high}$, respectively.
The more transferable one is bolded.}
\centering
\resizebox{0.8\linewidth}{!}{
\begin{tabular}{l|cccccccccccc|cc}

\hline
%\multirow{3}{c}{Measurement}&\multicolumn{12}{c}{Office-31}&\multicolumn{2}{c}{VisDA}\\

\multirow{3}{*}{Measurement}&\multicolumn{2}{c}{A$\rightarrow$D}&\multicolumn{2}{c}{A$\rightarrow$W}&\multicolumn{2}{c}{D$\rightarrow$A}&\multicolumn{2}{c}{D$\rightarrow$W}&\multicolumn{2}{c}{W$\rightarrow$A}&\multicolumn{2}{c}{W$\rightarrow$D}\vline& \multicolumn{2}{c}{Synthetic$\rightarrow$Real}\\
%\cmidrule(r){2-3}\cmidrule(r){4-5}\cmidrule(r){6-7}\cmidrule(r){8-9}\cmidrule(r){10-11}\cmidrule(r){12-13}  \cmidrule(r){14-15}
& $Z_{low}$ & $Z_{high}$& $Z_{low}$ & $Z_{high}$& $Z_{low}$ & $Z_{high}$& $Z_{low}$ & $Z_{high}$& $Z_{low}$ & $Z_{high}$& $Z_{low}$ & $Z_{high}$&$Z_{low}$ & $Z_{high}$ \\
\hline
%source-only &44.6& 67.3 &74.0\\
%$\lambda_1$ &0.5/78.7 &2/79.5&5/81.2 &10/78.0 &15/77.2&25/75.7  \\
%$\lambda_2$ &0.03/75.8 &0.1/77.7&0.3/80.4 &1/81.2 &3/80.4&10/68.6  \\

MMD$\downarrow$&\textbf{0.95}&0.96&\textbf{0.23}&0.26&\textbf{0.50}&0.55 &\textbf{0.50}&0.53   &\textbf{0.16}&0.17  &0.30&\textbf{0.27}   &\textbf{0.55}&0.61 \\
A-Distance$\downarrow$&\textbf{1.72}&1.81&\textbf{1.70}&1.78&\textbf{1.64}&1.77    &\textbf{ 0.93}&1.35 &\textbf{1.42}&1.45  &\textbf{0.93}&1.2   &\textbf{0.16}&0.17\\
Corresponding Angle$\uparrow$ &\textbf{0.7}&0.61&\textbf{0.15}&0.11&\textbf{0.10}&-0.05  &\textbf{0.57}&0.12  &\textbf{0.27}&-0.02 &\textbf{-0.2} &-0.3   &\textbf{0.38}&0.11\\
LogME$\uparrow$&\textbf{0.74}&0.70&\textbf{0.75}&0.72&\textbf{0.61}&0.60&\textbf{0.74}&0.63 &\textbf{0.64}&0.63    &\textbf{0.78}&0.76  &\textbf{0.21}&0.20 \\
LEEP$\uparrow$&\textbf{-2.51}&-2.65&\textbf{-2.11}&-2.41&\textbf{-3.01}&-3.55  &\textbf{-2.44}&-2.51 &-3.14&\textbf{-3.00}  &\textbf{-2.00}&-2.01   &\textbf{-0.20}&-0.22\\
NCE$\uparrow$&\textbf{-0.55}&-0.79&\textbf{-0.69}&-0.84&\textbf{-1.64}&-1.55&\textbf{-0.28}&-0.38   &\textbf{-1.51}&-1.65&\textbf{-0.14}  &-0.15      &-1.03&\textbf{-0.32}  \\
Accuracy(\%)$\uparrow$&\textbf{80.3}&73.1&\textbf{74.8}&68.4&\textbf{54.6}&50.0&\textbf{92.5}&89.1   &\textbf{59.6}&55.4&\textbf{98.7}  &97.1     &\textbf{51.3}&45.9  \\
%Accuracy$\uparrow$&\textbf{45.5\%}&42.2\%&\textbf{66.4\%}&64.2\%&\textbf{73.9\%}&70.5\%\\
%$\lambda_1$ &0.03/79.5 &0.05/80.8&0.08/81.2 &0.12/81.0 &0.2/78.8&0.3/78.6  \\
%$\lambda_2$ &0.03/79.5 &0.05/80.8&0.08/81.2 &0.12/81.0 &0.2/78.8&0.3/78.6  \\
\hline
\end{tabular}}

\label{tab_transferability2}

\end{table*}

\vspace{-2mm}
\subsection{Empirical Analysis of UTR}
\vspace{-1mm}
\label{Further Empirical Analysis of UTR}

\noindent \textbf{The effectiveness of $UTR_D$.}
The $UTR_D$ describes the domain-level transferability over the channel axis, which identifies how transferable each channel of the source encoder is to the target domain.  To evaluate the effectiveness of UD, we conduct the following experiments.

\textit{Implementation.} The experiments are conducted on the Office-31, Office-Home, and VisDA tasks.
For the Office-31 tasks and the Office-Home tasks, the backbone of ResNet-50 along with a fc layer is the source encoder, whose output channel $d=256$.  A fc layer with  weight normalization is the classifier. 
For the VisDA tasks, we replace the ResNet-50 with the ResNet-101 and keep the other settings the same.
We follow \cite{liang2020we,yang2021exploiting} to train the source model. For each task, we feedforward all target data to the pre-trained source model and finally calculate  $UTR_D(h_s)$ according to Equation \ref{eqa:UTR_D}, where $T=2$, $r_t$ is randomly sampled from $U(-0.05,0.05)$.

\textit{Comparison Protocols.}
We evaluate our $UTR_D$ by comparing it with the existing transferability measurements, that are \textit{ inapplicable in the SFUDA}, including: MMD \cite{gretton2006kernel}, A-Distance \cite{ben2010theory}, Corresponding Angle \cite{chen2019transferability},  LogME \cite{you2021logme}, LEEP \cite{nguyen2020leep} and NCE\cite{tran2019transferability}. 
In addition, the performance (prediction accuracy) is also considered as an extra intuitive measurement.
Considering that existing transferability measurements are not suitable for a single channel's feature, we design the following comparison protocol. Specifically,
we sort  the 256 channels representations $z=h_s(x)$  and  split them into two separate 128 channels vectors $Z_{low}$ and $Z_{high}$, representing the channels with the 128 smallest $UTR_D^i(h_s)$ and the 128 largest $UTR_D^i(h_s)$ respectively. 
In other words, the conclusion of  $UTR_D(h_s)$ is that $Z_{low}$ is  more transferable than $Z_{high}$.
Then we calculate the existing transferability measurements on $Z_{low}$ and $Z_{high}$ and report the consistency of their conclusion with ours. Note that these source data and target annotation are given when using these measurements. The results are reported in Table \ref{tab_transferability1} and \ref{tab_transferability2}.

% We use a ResNet-50\cite{he2016deep} along with an extra fully connected (fc) layer (256-dim) to extract feature representation $z \in \mathbb{R}^{ 256}$, and a fc layer (65-dim, corresponding to the number of categories of Office-Home) with a weight normalization layer as classifier $g$.
% For each task, we train the source model, then apply it to target domain data and finally calculate the Equation \ref{eqa:$UTR_D$f} to get $$UTR_D$(z)=[$UTR_D$(z_1),$UTR_D$(z_2), ..., $UTR_D$(z_{256})]$.

First, we quantitatively measure the transferability of $Z_{low}$ and $Z_{high}$ with the two vanilla UDA methods, the MMD and the A-Distance, which requires the source data. 
These methods measure the ability to bridge the domain discrepancy between the source and target domain. The lower MMD/A-Distance, the more transferable the model is.
From Table \ref{tab_transferability1} and \ref{tab_transferability2}, it can be seen that in 15 out of 19 adaptation tasks, the MMD of $Z_{low}$ is lower than that of $Z_{high}$, and in 17 tasks, the A-Distance of $Z_{low}$ is lower than that of $Z_{high}$. The result indicates that $UTR_D$ is consistent with these domain discrepancy measurements in most case, which suggest features in channels with less $UTR_D$ are more effective to bridge the domain discrepancy, therefore; they are more transferable.

%Taking the MMD as an example, we first extract $Z_{low}$ features from all target data and source data using the pre-trained source model, denoted as $F_{t;low}$ and $F_{s;low}$, respectively. And similiarly, we get $Z_{high}$ dimensions' feature representations $F_{t;high}$ and $F_{s;high}$.
%Next we calculate the MMD following \cite{gretton2006kernel} between $F_{t;low}$ and $F_{s;low}$ (denoted as $MMD_{low}$), and between $F_{t;high}$ and $F_{s;high}$ (denoted as $MMD_{high}$)
%Then the $MMD_{low}$ and $MMD_{low}$ denote the MMD measurements to $Z_{low}$ features and $Z_{high}$ features, respectively. 
%With the same strategy, we can get the A-Distance to $Z_{low}$  and $Z_{high}$ ($A_{low}$ and $A_{high})$ and the corresponding angle ($Cor_{low}$ and $Cor_{high}$).

Second, we compared with the Corresponding Angle, which is proposed by Chen et al. \cite{chen2019transferability} according to their observation that the eigenvectors with the largest singular values will dominate the feature transferability. 
We can observe that in 17 cases, the Corresponding Angle of $Z_{low}$ is larger than that of $Z_{high}$ in all experiments, demonstrating the consistency of $UTR_D$ with the Corresponding Angle. 
% Note, take $Z_{low}$ as an example; given the extracted $z$, we only remain its values in $Z_{low}$ dimensions and set other dimensions to 0, and then forward it to the classifier to get a prediction. 

Third, we compare the consistency of our method with the transferability measurements LogME, NCE, and LEEP that estimate the potential of the  source model parameter in learning a well-performed target model by refining.
In Table \ref{tab_transferability1} and \ref{tab_transferability2}, it can be seen that $UTR_D$ is consistent with LogME in all tasks , and also consistent with NCE and LEEP in 18 and 17 cases, respectively. These results denote that the relevant source model parameters to encode $Z_{low}$ is more transferable than $Z_{high}$.

Finally, We also calculate the  classification performances of $Z_{low}$ and $Z_{high}$ on the target domain.
Using $Z_{low}$, for example, we set the features of channels which \textit{not belongs to} $Z_{low}$ to zero, feedforward the modified feature into the classifier to get prediction and calculate the prediction accuracy..
$Z_{high}$ is evaluated in the same way.
% Note, take $Z_{low}$ as an example, given the extracted $z$, we only remain its values in $Z_{low}$ dimensions and set other dimensions to 0, and then forward it to the classifier to get a prediction. 
As shown in Table \ref{tab_transferability1} and \ref{tab_transferability2}, the prediction using $Z_{low}$ is more accurate than $Z_{high}$ in target domain on  all  adaptation tasks. 
For example, the accuracy of $Z_{low}$  outperform $Z_{high}$ by 3.1\%, 4.8\% and 1.3\% on Pr$\rightarrow$Ar, Pr$\rightarrow$Cl and Pr$\rightarrow$Re, respectively. Note that we did not extra train the source model but only split it into two part of channels $Z_{low}$ and $Z_{high}$ according to our $UTR_D$. The significant performance gap between the two parts
indicates that $Z_{low}$ with less $UTR_D$ is more transferable to the target domain than $Z_{high}$.
The experimental observations from the series of studies above illustrate that 1) the proposed $UTR_D$ is strongly consistent with current transferability measurements and can estimate the transferability, 
2) Our method can effectively analyse the internal transferability of the source model, the channels of source encoder with lower $UTR_D$  is  more transferable to the target domain, which allows us to leverage the source knowledge more efficiently and safely, which proves our motivation.

\begin{figure}[t]
  \centering

    \includegraphics[width=0.3\linewidth]{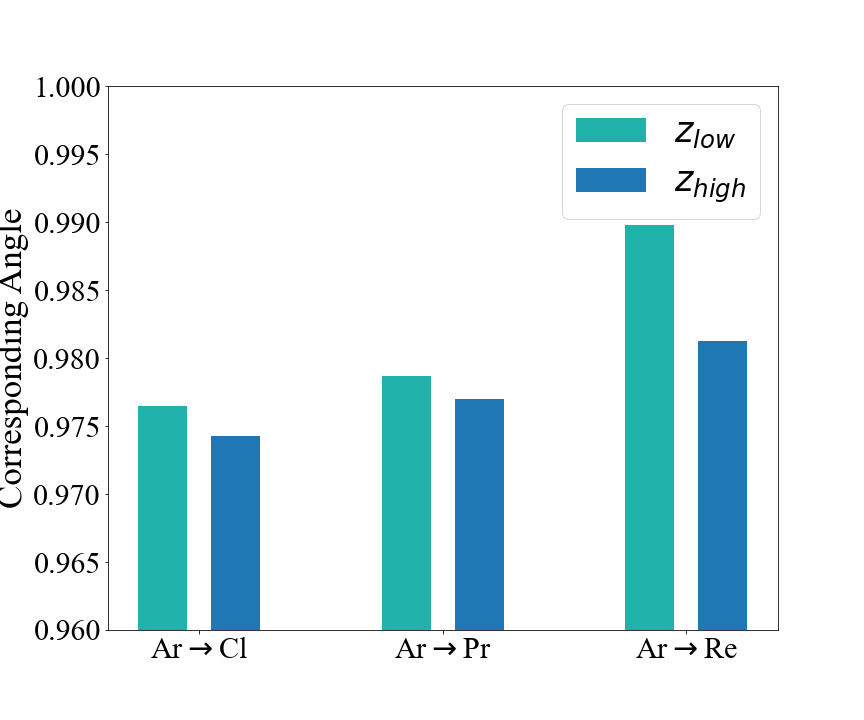}
    \includegraphics[width=0.3\linewidth]{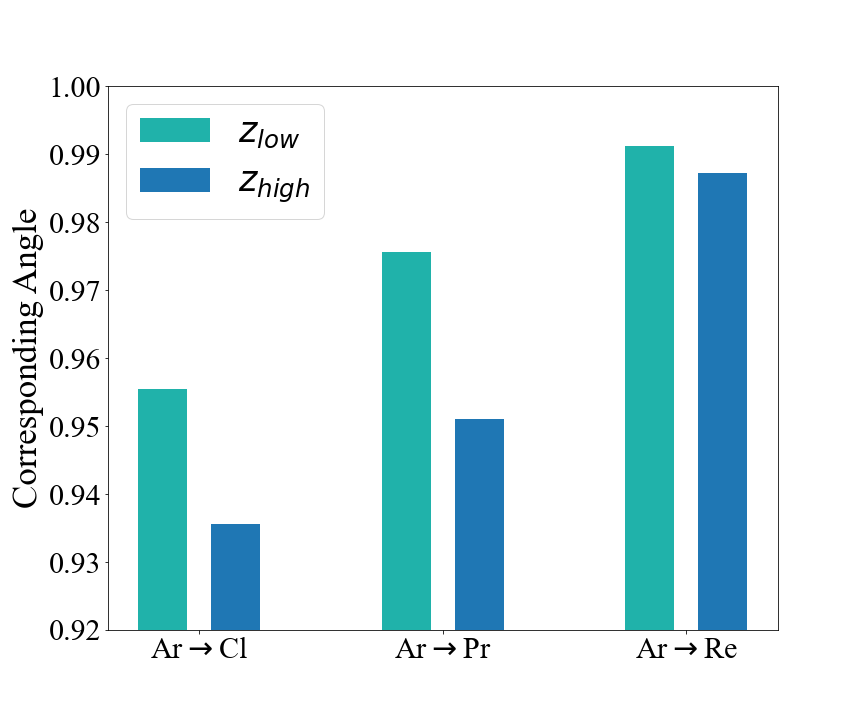}
    \includegraphics[width=0.3\linewidth]{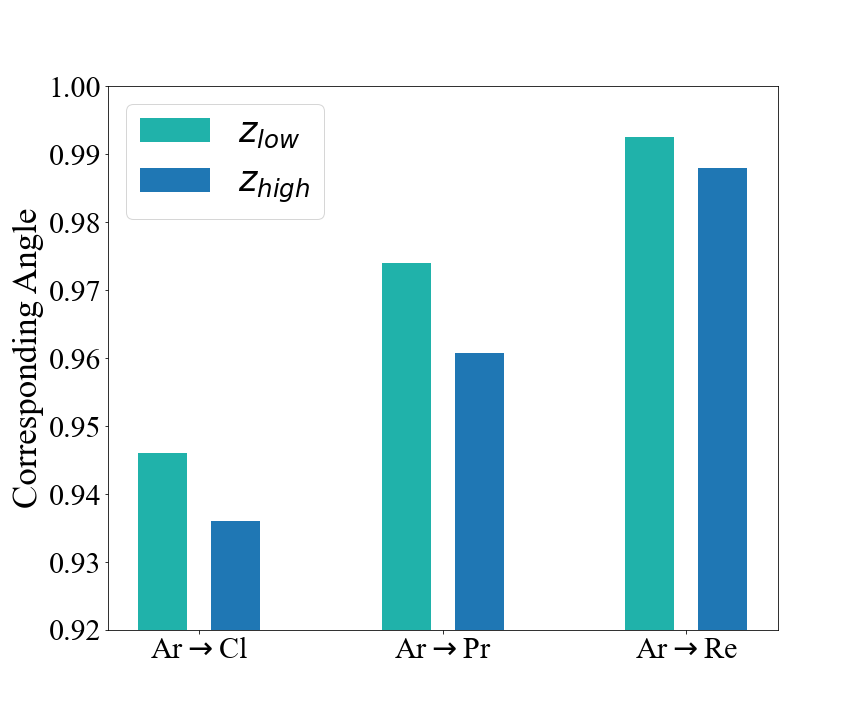}
    \includegraphics[width=0.3\linewidth]{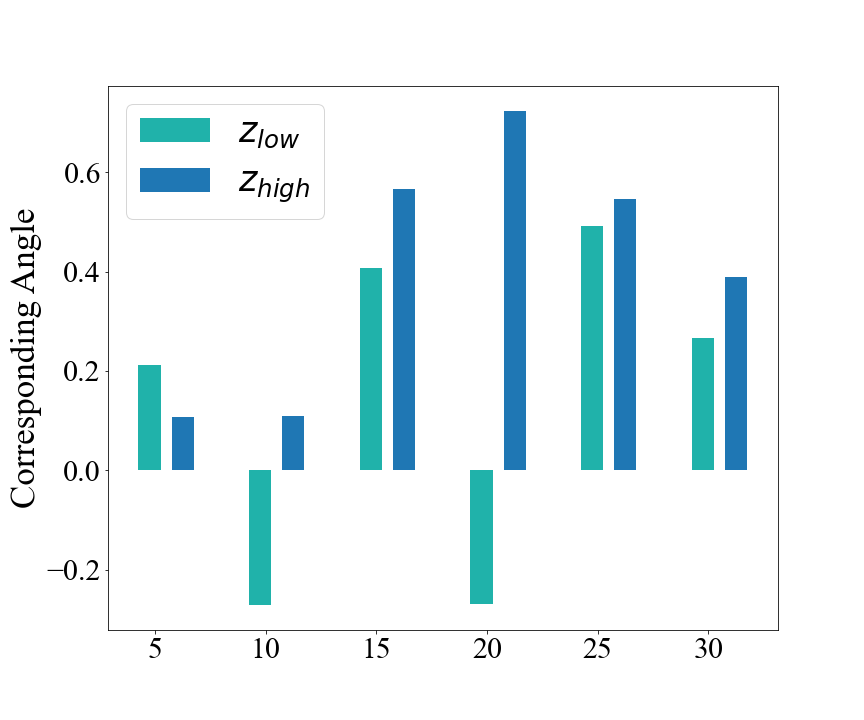}
    \includegraphics[width=0.3\linewidth]{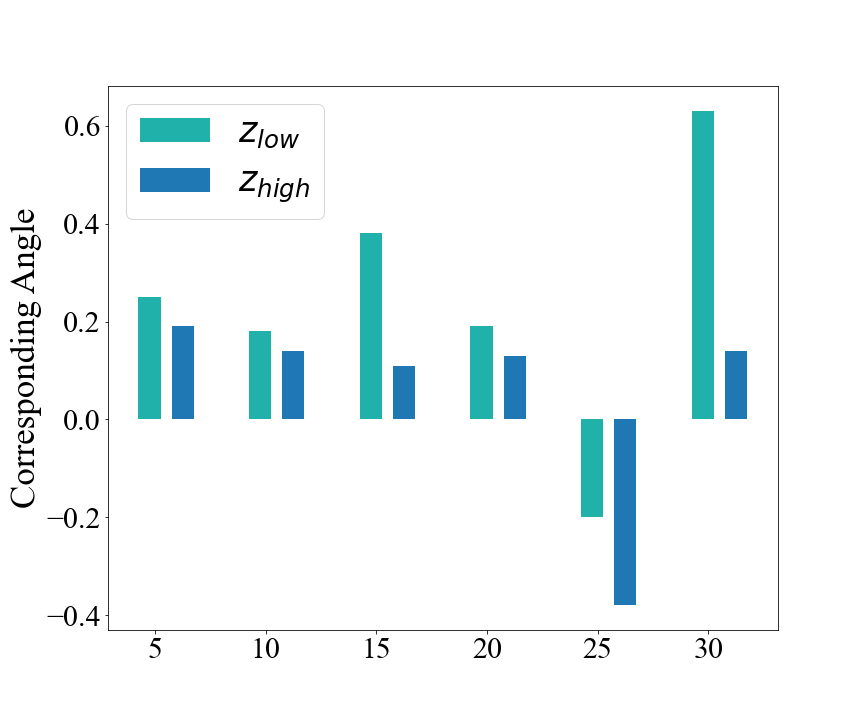}
    \includegraphics[width=0.3\linewidth]{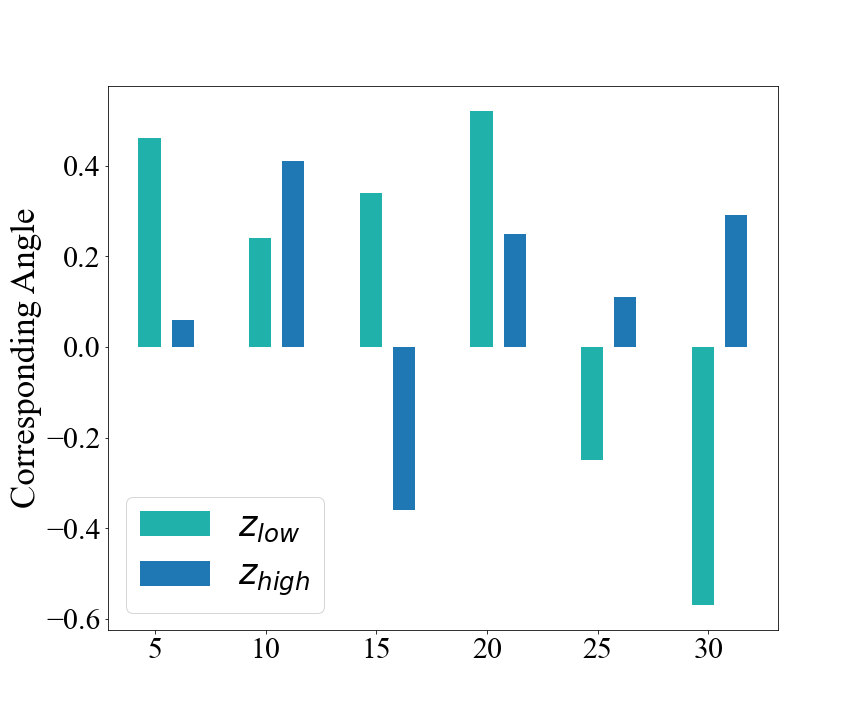}
    
    \vspace{-2mm}
   \caption{ (a)-(c): The effectiveness of the $UTR_D$ for channels of different layers. (a): At the last layer. (b) At the penultimate layer. (c) At the antepenultimate layer. (d)-(f): The effectiveness of the $UTR_D$ to the target model during the adaptation process (trained with 5, 10, 15, 20, 25 and 30 epochs) on (d) $Ar\rightarrow Cl$, (e) $Ar\rightarrow Pr$, and (f) $Ar\rightarrow Re$. $Z_{low}$ and $Z_{high}$   represent the channels with the 128 smallest $UTR_D^i(h_s)$ and the 128 largest $UTR_D^i(h_s)$ respectively.}
   \vspace{-3mm}
   \label{fig:corrlayer}
\end{figure}

\noindent \textbf{The effectiveness of $UTR_I$.}
 The $UTR_I$ identifies the instance-level reliability of inferring target semantics using the source model. To evaluate its effectiveness, we draw the  Accuracy-$UTR_I$ curve in Fig. \ref{fig:acc_UD_UTR}, which describes the relationship between the $UTR_I$ of different target samples and the prediction accuracy of the source model to these samples. It can be seen that the source model is more accurate to samples with small $UTR_I$. And the prediction accuracy tends to decrease with the increase of the $UTR_I$. This phenomenon demonstrates the  effectiveness of  $UTR_I$ to identify the instance-level risk of inferring target semantic labels.
 %In addition, it can be seen that $UTR_I$ is more efficient than UD. The reason is that the UD directly consider the uncertainty of the whole model (the feature encoder and the classifier) to the target sample, while the $UTR_I$ integrate the uncertainty of different features. Based on the theoretical results that the $g_{s}$ is closely related to the optimal target classifier  \cite{kuzborskij2013stability, liang2020we}, there is little uncertainty in the classifier, and the uncertainty caused by the domain discrepancy, is mainly contained in the feature encoder. As a result, $UTR_I$ can capture domain discrepancy more effectively, resulting in more accurate estimates of the transferability of the model to target samples, therefore better suggesting the reliability the inferred  target semantics.
%\noindent \textbf{The effectiveness of UD.}
%Our UTR explores and exploits the Uncertainty Distance (UD) to estimate the transferability of a source model to a target sample. To evaluate the effectiveness of UD, we draw the curve in Fig. \ref{fig:acc_UD_UTR} to describe the relationship between the UD value of the source model to different target samples and the prediction accuracy of the source model to these samples. It can be seen that the source model is more accurate to samples with small UD, suggesting the source model is  transferable to these samples. When the threshold of UD is increasing, the prediction accuracy tend to decrease, which denotes the source model is not transferable to these samples, which proved our motivation.

\noindent \textbf{Extension to other layers.} In our previous experiments, UTR is calculated using the last layer output of the feature extractor (the FC layer). Here, we explore the feasibility of extending the UTR to channels of other layers. To this end, we use the average pooling to extract features of different channels $z \in \mathbb{R}^ {2048}$ from the last, penultimate, and antepenultimate bottleneck of the ResNet-50 backbone, respectively. 
Then we calculate the UTR  of these channels and evaluate the effectiveness of $UTR_D$ and $UTR_I$. For the $UTR_D$, the evaluation method is similar to the previous one: that is, the feature $z$ is divided into two 1024 channels vectors  $Z_{low}$ and $Z_{high}$ according to $UTR_D(h_s)$, and their corresponding angles are compared to evaluate their transferability. The results are shown in Fig. \ref{fig:corrlayer}.
We can see that the corresponding angle of $Z_{low}$ with lower $UTR_D$, is larger than $Z_{high}$ with higher $UTR_D$. Therefore, the $UTR_D$ is effective at the last, penultimate, and antepenultimate bottlenecks of Resnet-50 as well. The similar trends among multiple layers' features prove that our transferable index has the potential to extend to the feature representation of other layers.
 For the $UTR_I$, the Accuracy-$UTR_I$ curve is shown in Fig. \ref{fig:acc_other_UTR} (a). It can be seen that the $UTR_I$ is satisfying in the last bottleneck of the ResNet-50 backbone. On the penultimate, and antepenultimate bottlenecks, it may be invalid,  such as when $\tau=2.5$ but is effective overall.

\begin{figure}[t]
\vspace{-2mm}
\centering
\includegraphics[width=0.9\linewidth]{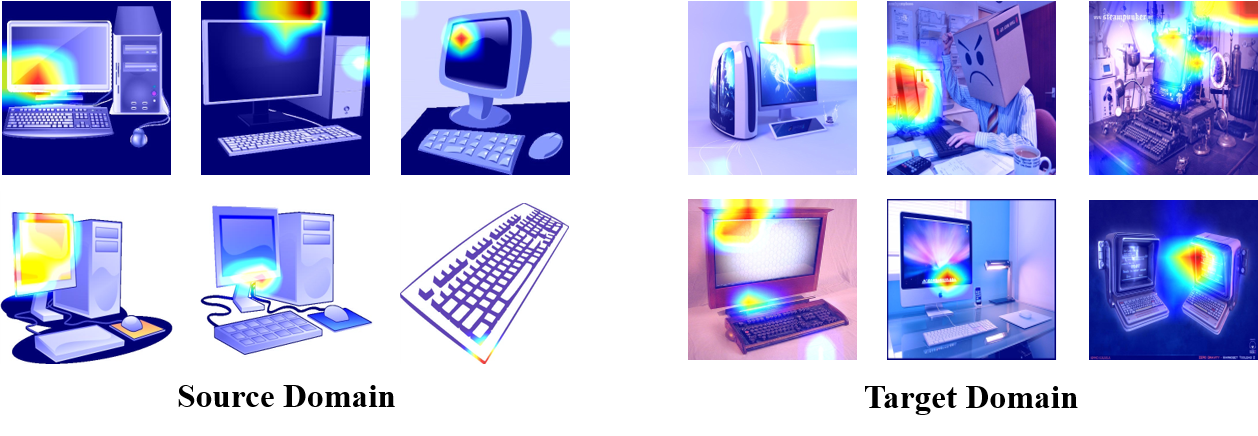}
\caption{The Grad-CAM \cite{selvaraju2017grad} visualization of the features of channels with the smallest $UTR_D$  on the source and target domains. }
\label{fig_smallest}
\end{figure}

\begin{figure}[t]
\vspace{-2mm}
\centering
\includegraphics[width=0.9\linewidth]{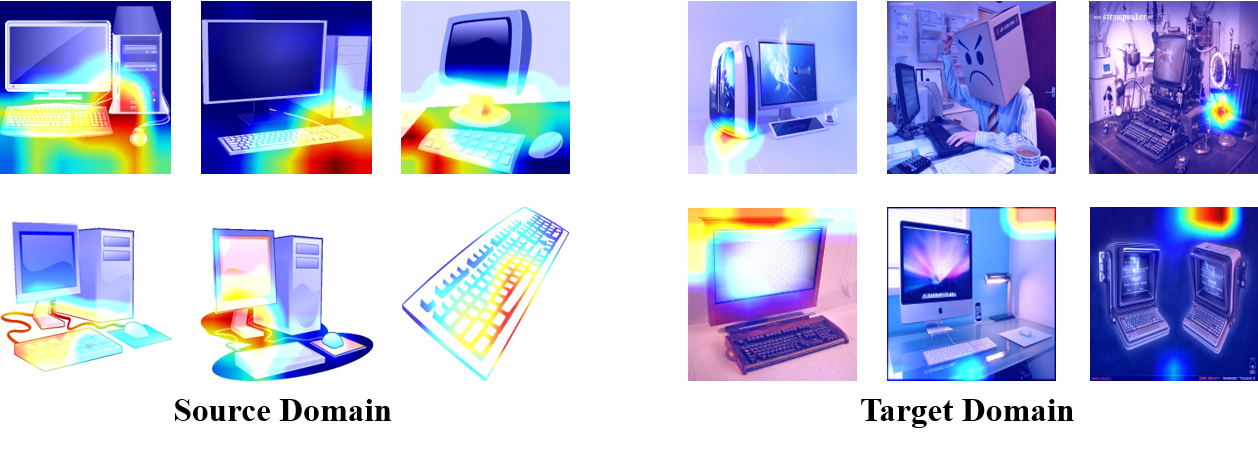}
\caption{The Grad-CAM \cite{selvaraju2017grad} visualization of the features of channels with the largest  $UTR_D$  (b)  on the source and target domains. }

\label{fig_largest}
\end{figure}

\begin{figure}
  \centering
 
\includegraphics[width=0.4\linewidth]{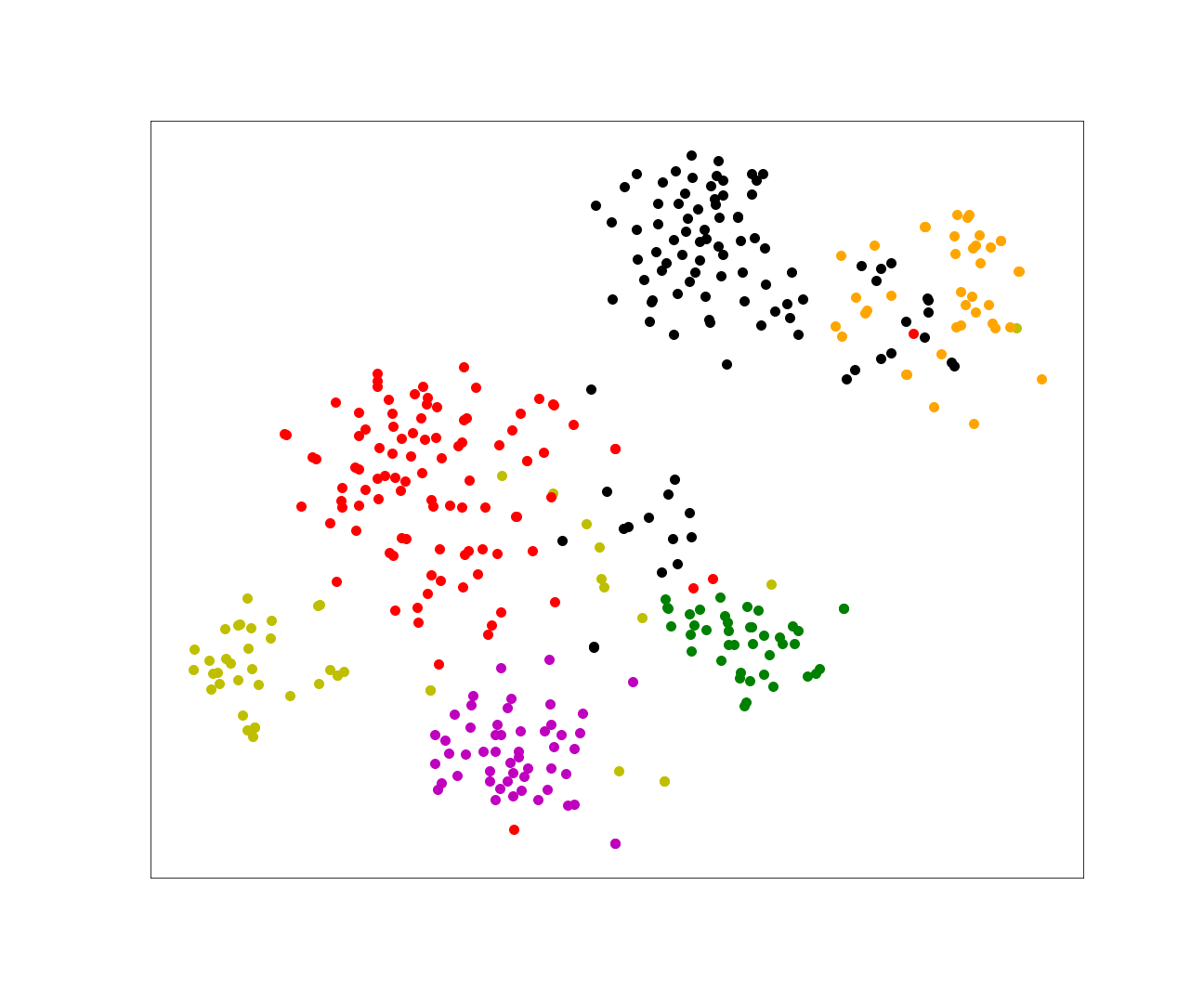}
\includegraphics[width=0.4\linewidth]{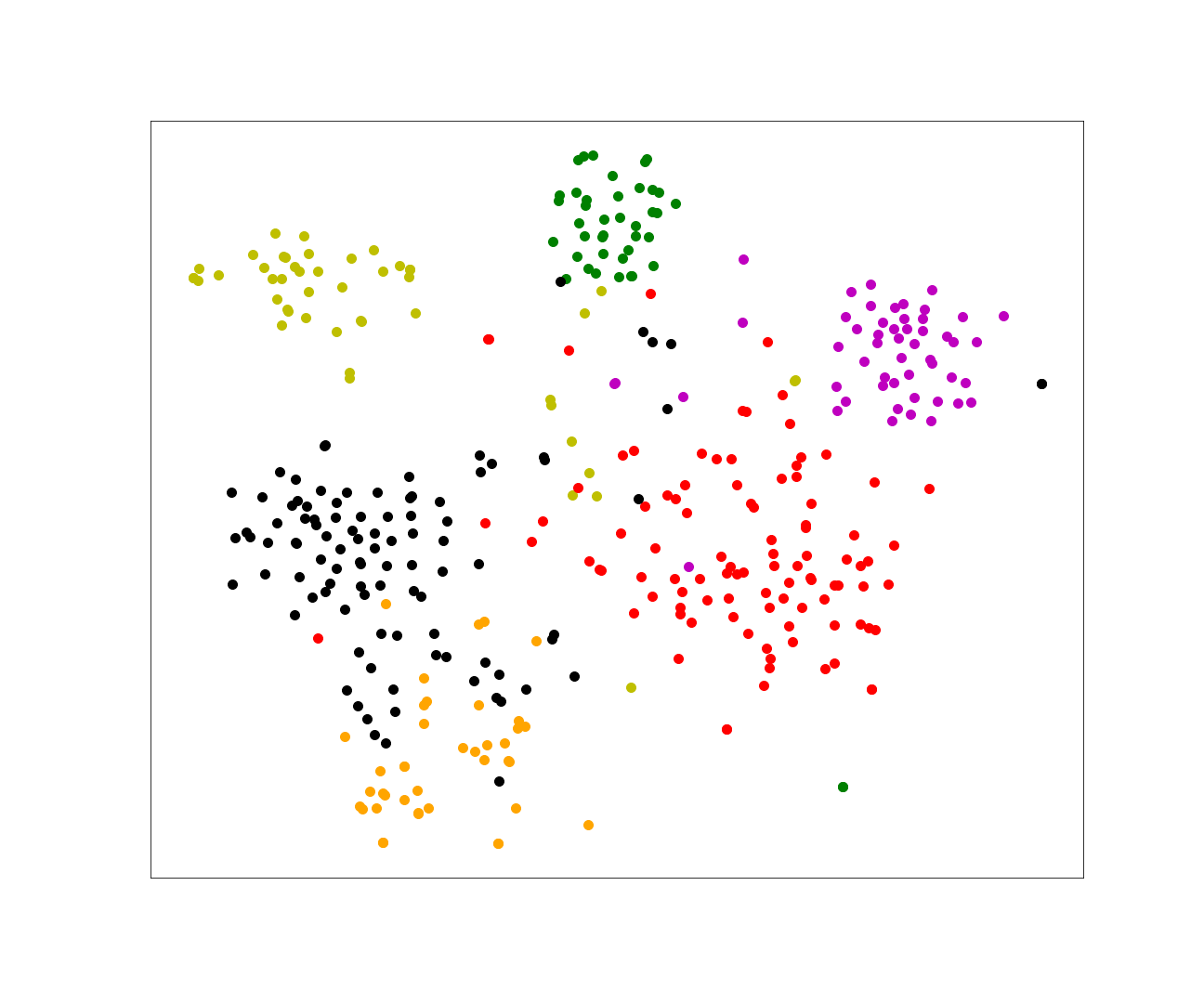}
\vspace{-1mm}
\caption{The t-SNE visualizations of features of $Z_{low}$ and $Z_{high}$ on the target domain (Cl) of the Office-Home task Ar$\rightarrow$Cl. For a better illustration, we choose features in the first 6 classes, and different color denotes different class. Best viewed in colors.}
\label{fig:z_tSNE}
\end{figure}

\begin{figure}
  \centering

\includegraphics[width=0.4\linewidth]{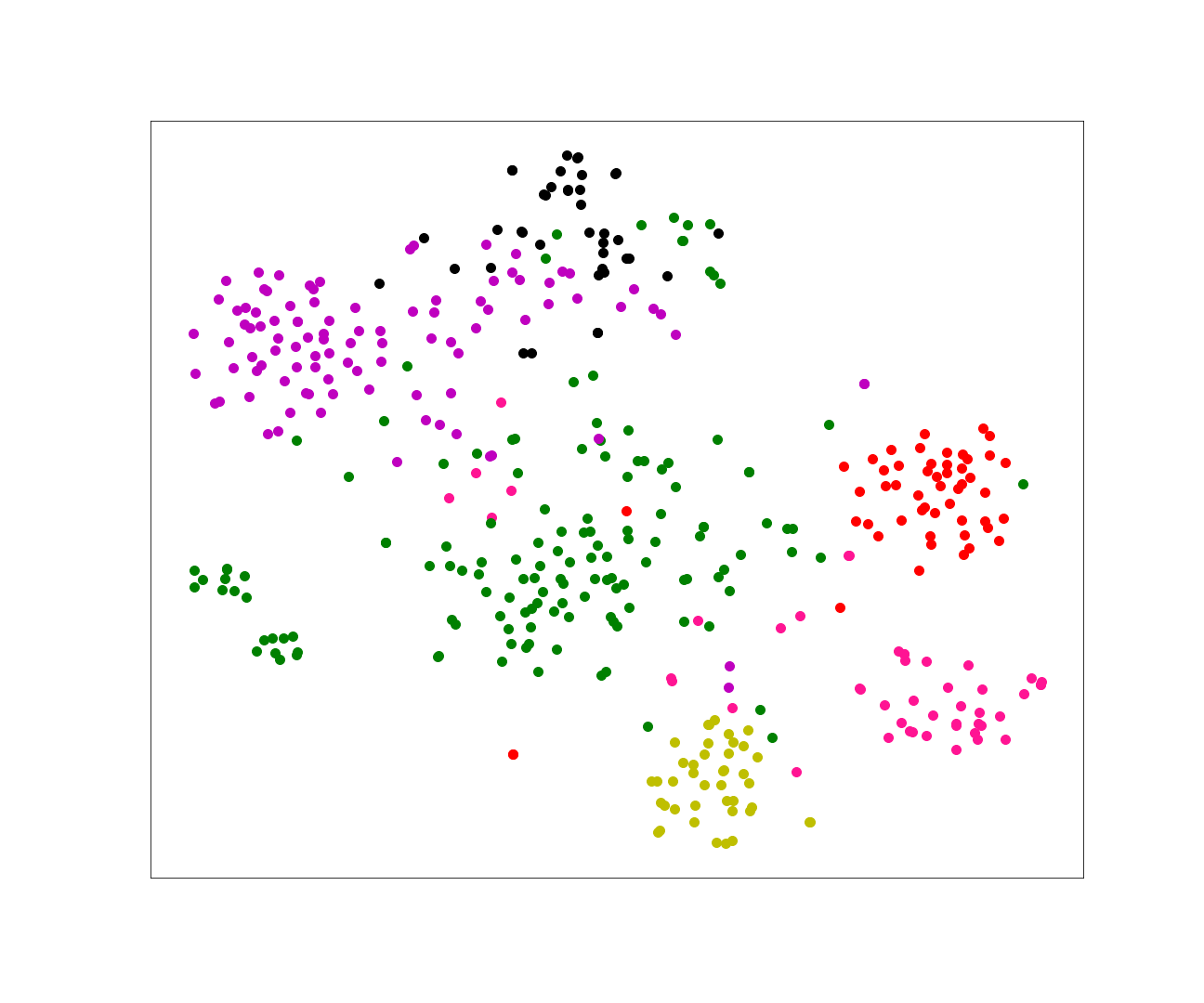}
\includegraphics[width=0.4\linewidth]{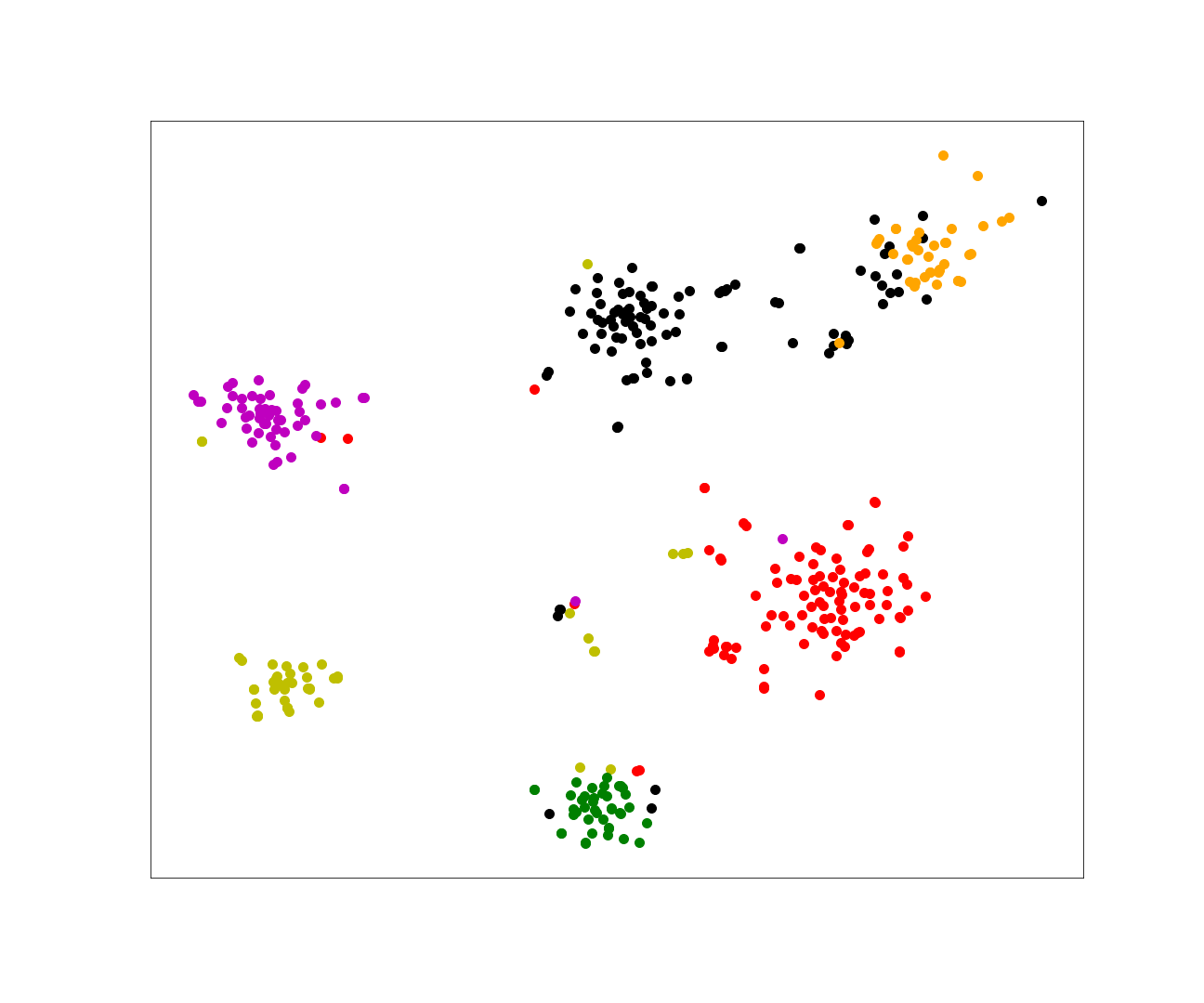}

\includegraphics[width=0.4\linewidth]{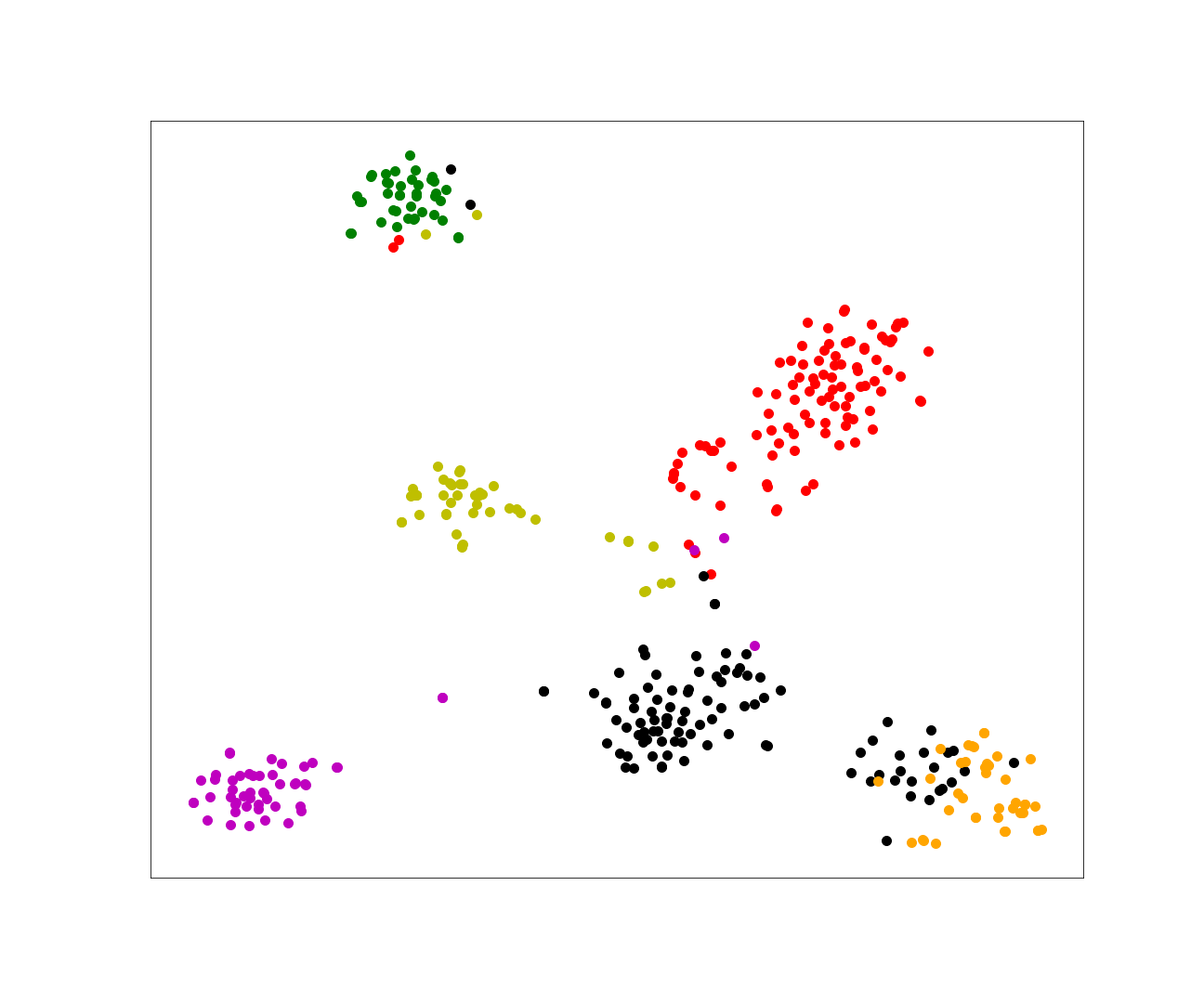}
\includegraphics[width=0.4\linewidth]{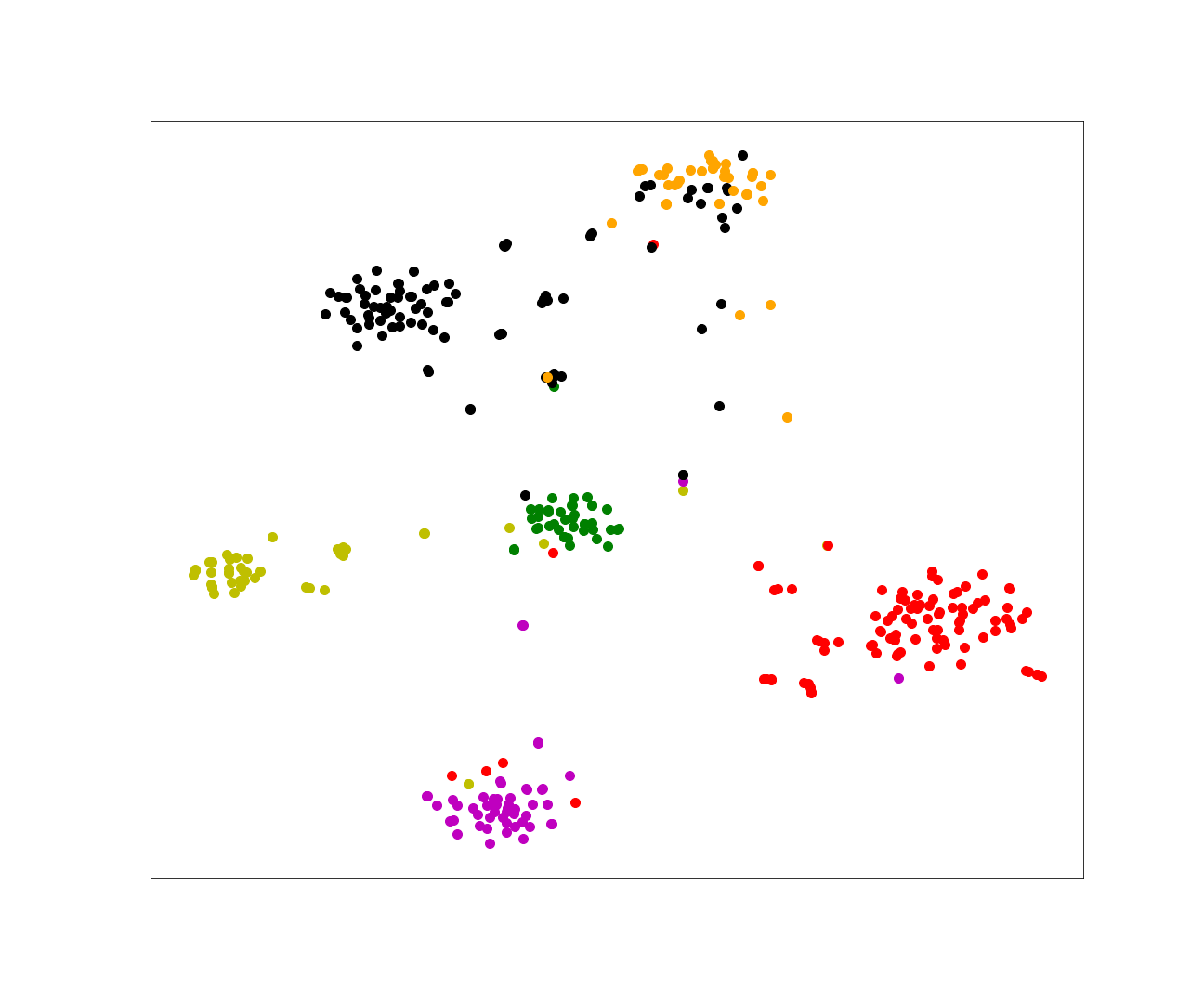}

\vspace{-3mm}
\caption{The t-SNE visualizations of different methods on the target domain (Ar) of the Office-Home task Cl$\rightarrow$Ar, including: the source model, Shot, NRC and Ours. For a better illustration, we choose features in the first 6 classes, and different color denotes different class. Best viewed in colors.}
\label{fig:tSNE}

\end{figure}
\begin{table}

\caption{Comparison with existing transferability measurements with different model structures on  office-home tasks Ar$\rightarrow$Cl. The $\uparrow$/$\downarrow$ indicates the larger/smaller the value, the higher the transferability.}
\centering
\resizebox{\linewidth}{!}{
\begin{tabular}{lcccccc}

\hline
Measurement&MMD$\downarrow$&A-Distance$\downarrow$&CA$\uparrow$&LogME$\uparrow$&LEEP$\uparrow$&NCE$\uparrow$\\
%\cmidrule(r){2-3}\cmidrule(r){4-5}\cmidrule(r){6-7}
\hline
%source-only &44.6& 67.3 &74.0\\
%$\lambda_1$ &0.5/78.7 &2/79.5&5/81.2 &10/78.0 &15/77.2&25/75.7  \\
%$\lambda_2$ &0.03/75.8 &0.1/77.7&0.3/80.4 &1/81.2 &3/80.4&10/68.6  \\

VGG16 $Z_{low}$&\textbf{0.60}&\textbf{1.36}&\textbf{-0.01}&\textbf{0.81}&\textbf{-3.61}&\textbf{-2.36}\\
VGG16 $Z_{high}$&0.64&1.44&-0.21&0.80&-3.63&-2.43\\
\hline
AlexNet $Z_{low}$&\textbf{0.51}&1.31&\textbf{0.50}&\textbf{0.76}&\textbf{-3.80}&\textbf{-2.88}\\
AlexNet $Z_{high}$&0.57&\textbf{1.28}&0.25&0.75&-3.85&-2.97\\
\hline

%Accuracy$\uparrow$&\textbf{45.5\%}&42.2\%&\textbf{66.4\%}&64.2\%&\textbf{73.9\%}&70.5\%\\
%$\lambda_1$ &0.03/79.5 &0.05/80.8&0.08/81.2 &0.12/81.0 &0.2/78.8&0.3/78.6  \\
%$\lambda_2$ &0.03/79.5 &0.05/80.8&0.08/81.2 &0.12/81.0 &0.2/78.8&0.3/78.6  \\
\end{tabular}}

\label{tab_transferability_models}

\end{table}

\noindent \textbf{Extension to other network architectures.} 
In this section, we evaluate the effectiveness of UTR on different backbone models including the VGG16\cite{simonyan2014very} and AlexNet\cite{krizhevsky2012imagenet}. The experiments are conducted on Office-Home task Ar$\rightarrow$Cl. The results of $UTR_D$ are reported in Table \ref{tab_transferability_models}.
The results of $UTR_I$ is shown in Fig. \ref{fig:acc_other_UTR} (b).
It can be seen that using two different backbone architectures, the $UTR_D$ is consistent with the most recent transferability measurements. In addition, the $UTR_I$ is able to reveal the target semantics risk, which demonstrates that our UTR method is able to apply to different model architectures.

\noindent \textbf{Extension to the target model.} 
We have investigated the effectiveness of our UTR on the source model. In this subsection, we evaluate it on the target model in the adaptation process. 
The results of $UTR_D$ and $UTR_I$  are shown in Fig. \ref{fig:corrlayer} (d)-(f) and \ref{fig:acc_other_UTR} (c), respectively.
From Fig. \ref{fig:corrlayer} (d)-(f), we can observe that the $UTR_D$ is also effective for the target model in the first few steps of adaptation. 

To be specific, we can see that in the first 5 steps, the $Z_{low}$ has a larger Corresponding Angle between the source and the target domain than $Z_{high}$, which indicates that it is more transferable than $Z_{high}$.
However, it can be seen that after  training for a period, it is inadequate to use the $UTR_D$ for identifying the target model. For example, in the epoch 10/30 of Fig. \ref{fig:corrlayer} (d), the corresponding angle of $Z_{low}$ is lower than that of $Z_{high}$. 

The same phenomenon can be observed for $UTR_I$. From Fig. \ref{fig:acc_other_UTR} (c), we can observe that the $UTR_I$ is effective in the epoch 0 and epoch 5, but became invalid in the epoch 15.
The main reason may be that after a period of training, the model gradually adapts to the target domain. 
Thus, it no longer needs or even actively abandons the source knowledge. It is worth noting that if the target model does not fit the source domain well, the model uncertainty in the Equation \ref{eqa:unc_cal_1} can not be ignored. Therefore the UD may not be calculated without accessing to the source data.
%This is why we terminate the knowledge absorption $L_{kd}$ after 10 epoch by setting $\lambda =0$.

%\begin{figure}
%  \centering
 
 %   \includegraphics[width=1.0\linewidth]{figs/num-acc.pdf}
%    \vspace{-4mm}
%   \caption{The accuracy-number curve of high risk sample with changing threshold $\tau$  on three task: (a) Ar$\rightarrow$Cl , (b) Ar$\rightarrow$Pr, and (c) Ar$\rightarrow$Re. The blue line denotes the prediction accuracy on the selected high-risk samples, and the red line corresponds to the number of them.   }
 %  \label{fig:num-acc}

%\end{figure}

\noindent \textbf{Calculating the $UTR_D$ stochastically.} 
In our implementation, the calculation of $UTR_D$ requires to feed-forward all target samples. As a statistical measurement, $UTR_D$ can also be adapted to the online version, where $UTR_D$ is updated using the moving average method widely used in Batch Normalization\cite{ioffe2015batch}. We conducted new experiments using the moving-average calculation of $UTR_D$, with their results denoted as ``Ours-Online''. We set the momentum to 0.1 and conduct the experiment on three office-home tasks: Ar$\rightarrow$Cl, Ar$\rightarrow$Pr, and Ar$\rightarrow$Re. The results in Table \ref{tab_abl_module} show that the performances of the online version on the three tasks are 59.2\%, 81.2\% and 83.1\%, respectively. These are very similar to the original version ``Ours''.

\noindent\textbf{Uncertainty Estimation.} We evaluate the performances of using different uncertainty implementation  methods to calculate $UTR_D$, i.e., the $M(.)$ in Equation \ref{eqa:mUTR_D}, including the sensitivity analysis \cite{nagy2007distributional} and Deep Ensembles \cite{lakshminarayanan2016simple} and Monte Carlo dropout \cite{gal2016dropout}, denoted as ``Ours-Ensemble'' and ``Ours-MC drooout'', respectively. Table \ref{tab_abl_module} shows that our method is not sensitive to various uncertainty implementation methods.

\noindent\textbf{Visualization.} 
Fig. \ref{fig_smallest} and Fig. \ref{fig_largest} illustrate the Grad-CAM \cite{selvaraju2017grad} feature visualization of a source model on the source and target data. Fig. \ref{fig_smallest} visualizes the feature of the most transferable channel selected by our proposed $UTR_D$ (i.e., with the smallest $UTR_D$). Fig. \ref{fig_smallest} shows the feature of the most non-transferable one (i.e., with the largest $UTR_D$). It can be observed that the feature in Fig. \ref{fig_smallest} captures the  semantic information “screen” on the source domain and it remains the same semantic information on the target domain, which indicates that it is transferable to the target domain. However, the feature in Fig. \ref{fig_smallest} seems to focus on “keyboard” in the source domain, but fails to capture the same semantics on the target data, which suggests it is non-transferable to the target domain.

Fig. \ref{fig:z_tSNE} shows the t-SNE \cite{van2008visualizing} visualizations of the features of $Z_{low}$ (128 channels' features with low $UTR_D$) and $Z_{high}$ (128 channels' features with high $UTR_D$)  on the task Cl$\rightarrow$Al.
We can see that the semantic information extracted by $Z_{low}$ in the target domain is more discriminative than $Z_{high}$. It qualitatively proves that it is more transferable to the target domain. The above phenomenons demonstrate that $UTR_D$ is effective to estimate the transferability of the knowledge in the source encoder.

By estimating the transferability of different channels of the source encoder, our method can incorporate more valuable knowledge into the target domain to learn a more discriminative target model. To prove it, we provide the t-SNE visualizations of the feature obtained by the original source model, SHOT, NRC and our method on the task Cl$\rightarrow$Al in Fig. \ref{fig:tSNE}. As expected, the feature extracted by our method is more semantically discriminative.

\vspace{-4mm}
\section{Method Limitation}
\vspace{-2mm}
In this paper, we propose the Uncertainty-induced Transferability Representation (UTR) to explore the transferability of the source model in the absence of source data and target annotations. We prove the effectiveness and universality of the domain-level UTR and the instance-level UTR, which help the SFUDA community leverage the knowledge of the source model and target data fully and safely. 
However, it also has the following two limitations.

First, we use the distributional uncertainty to approximate the implicit uncertainty distance, which assumes that the model uncertainty is small enough to be ignored. As we discussed in Section \ref{Further Empirical Analysis of UTR}: ``Extension to the target model'', because the model uncertainty represents how much the pre-trained model covers the training distribution, the assumption may be violated somehow with the model gradually adapted to the target domain. In this paper, it will be our future work to quantify when the previous assumption is violated.

Second, we demonstrate the consistency of $UTR_D$ with existing domain discrepancy measurements. However, at present it is only a way to analyse the transferability, but not a rigorous domain distribution divergence yet that can be explicitly optimized, such as MMD and A-Distance. We hope that future research will address this limitation.

% First, we use the distributional uncertainty to measure the implicit uncertainty distance, which require assuming the model uncertainty, the degree to which the fitted region of the model covers the training distribution, is small to be ignored. As the model gradually adapt to the target domain, this assumption becomes invalid, which limits the performance of the method.
% Second, we demonstrate the consistency of $UTR_D$ with existing domain discrepancy measurements. However, at present it is only a way to analyse transferability, but not the distance that can be explicitly optimized to reduce the domain discrepancy such as MMD and A-Distance.

%\input{samples/supplement}
%\noindent\underline{\textbf{Method Limitation and Social Impact}}
%In this paper, only the closed-set SFUDA problems are discussed but not the open-set SFUDA problems. Our SFUDA method has important implications for applications where the annotation is costly and with restricted data sharing policy.
\vspace{-4mm}
\section{Conclusions}
\vspace{-2mm}
In this paper, we develop a novel measurement termed Uncertainty-induced Transferability Representation (UTR) which uses uncertainty distance as a tool to estimate transferability in the absence of source data and target annotations. The  domain-level UTR describes how transferable each source feature dimension is to the target domain, and  the instance-level UTR identifies the reliability of the inferred target semantics.
Based on the UTR, we propose a novel Calibrated Adaption Framework (CAF) for SFUDA, including a source knowledge calibration module to control the target model to learn transferable knowledge and discard non-transferable one, and a target semantics calibration module  calibrates the target semantics.
The calibrated source knowledge and target semantics help the target model fully and safely leverage the source knowledge and target data, ultimately prompting to  better adapt to the target domain. We verified the effectiveness of our method using experimental results and demonstrated that the proposed method achieves state-of-the-art performances on three SFUDA benchmarks. 
%%%%%%%%% REFERENCES
\vspace{-3mm}
\bibliographystyle{IEEEtran}
\bibliography{ref.bib}

\vfill

\end{document}